\documentclass[11pt,a4paper]{article}
\usepackage[margin=1in]{geometry}
\usepackage{cite}
\usepackage{amsmath,amssymb}
\usepackage[table]{xcolor}
\usepackage{graphicx}
\usepackage{booktabs}
\usepackage{tabularx}
\usepackage{url}
\usepackage[hidelinks]{hyperref}
\usepackage{lmodern}
\usepackage[nopatch=footnote]{microtype}
\usepackage{multirow}
\usepackage{setspace}
\usepackage{listings}
\usepackage{array}
\usepackage{caption}
\usepackage{etoolbox}
\usepackage{titlesec}
\usepackage{ragged2e}
\usepackage{float}
\usepackage{needspace}
\usepackage[most]{tcolorbox}
\tcbuselibrary{listings,skins}

\definecolor{StageIHeader}{HTML}{24425B}
\definecolor{StageIStripe}{HTML}{F0F5FA}
\definecolor{StageIRule}{HTML}{24425B}

\definecolor{ArchitectureHeader}{HTML}{20665E}
\definecolor{ArchitectureStripe}{HTML}{F0F7F5}
\definecolor{ArchitectureRule}{HTML}{20665E}

\definecolor{EvaluationHeader}{HTML}{344C78}
\definecolor{EvaluationStripe}{HTML}{F1F4FA}
\definecolor{EvaluationRule}{HTML}{344C78}

\definecolor{AblationHeader}{HTML}{665078}
\definecolor{AblationStripe}{HTML}{F6F2F8}
\definecolor{AblationRule}{HTML}{665078}

\definecolor{ComparisonHeader}{HTML}{795A32}
\definecolor{ComparisonStripe}{HTML}{FAF6EE}
\definecolor{ComparisonRule}{HTML}{795A32}

\definecolor{AppendixAHeader}{HTML}{24425B}
\definecolor{AppendixAStripe}{HTML}{F0F5FA}
\definecolor{AppendixARule}{HTML}{24425B}

\definecolor{AppendixBHeader}{HTML}{20665E}
\definecolor{AppendixBStripe}{HTML}{F0F7F5}
\definecolor{AppendixBRule}{HTML}{20665E}

\definecolor{AppendixCHeader}{HTML}{665078}
\definecolor{AppendixCStripe}{HTML}{F6F2F8}
\definecolor{AppendixCRule}{HTML}{665078}

\definecolor{AppendixDHeader}{HTML}{795A32}
\definecolor{AppendixDStripe}{HTML}{FAF6EE}
\definecolor{AppendixDRule}{HTML}{795A32}

\definecolor{PromptBack}{RGB}{247,249,253}
\definecolor{PromptFrame}{RGB}{92,111,145}
\definecolor{OutputBack}{RGB}{246,251,248}
\definecolor{OutputFrame}{RGB}{76,125,102}
\definecolor{CodeBack}{RGB}{252,249,244}
\definecolor{CodeFrame}{RGB}{143,112,62}
\definecolor{CaptionGray}{RGB}{55,61,68}

\newcolumntype{Y}{>{\RaggedRight\arraybackslash}X}
\newcolumntype{Z}{>{\centering\arraybackslash}X}
\newcolumntype{L}[1]{>{\RaggedRight\arraybackslash}m{#1}}
\newcolumntype{C}[1]{>{\centering\arraybackslash}m{#1}}

\newcommand{\StageITableStyle}{\rowcolors{2}{StageIStripe}{white}\arrayrulecolor{StageIRule}}
\newcommand{\ArchitectureTableStyle}{\rowcolors{2}{ArchitectureStripe}{white}\arrayrulecolor{ArchitectureRule}}
\newcommand{\EvaluationTableStyle}{\rowcolors{2}{EvaluationStripe}{white}\arrayrulecolor{EvaluationRule}}
\newcommand{\AblationTableStyle}{\rowcolors{2}{AblationStripe}{white}\arrayrulecolor{AblationRule}}
\newcommand{\ComparisonTableStyle}{\rowcolors{2}{ComparisonStripe}{white}\arrayrulecolor{ComparisonRule}}
\newcommand{\AppendixATableStyle}{\rowcolors{2}{AppendixAStripe}{white}\arrayrulecolor{AppendixARule}}
\newcommand{\AppendixBTableStyle}{\rowcolors{2}{AppendixBStripe}{white}\arrayrulecolor{AppendixBRule}}
\newcommand{\AppendixCTableStyle}{\rowcolors{2}{AppendixCStripe}{white}\arrayrulecolor{AppendixCRule}}
\newcommand{\AppendixDTableStyle}{\rowcolors{2}{AppendixDStripe}{white}\arrayrulecolor{AppendixDRule}}

\AtBeginEnvironment{table}{%
  \sffamily%
  \setlength{\tabcolsep}{5.0pt}%
}

\newtcblisting{promptbox}{
  listing only,
  enhanced,
  colback=PromptBack,
  colframe=PromptFrame,
  boxrule=0.55pt,
  arc=1.5mm,
  left=1.5mm,right=1.5mm,top=1.1mm,bottom=1.1mm,
  before skip=5pt,after skip=6pt,
  listing options={basicstyle=\ttfamily\scriptsize,breaklines=true,columns=fullflexible,keepspaces=true,showstringspaces=false}
}
\newtcblisting{outputbox}{
  listing only,
  enhanced,
  colback=OutputBack,
  colframe=OutputFrame,
  boxrule=0.55pt,
  arc=1.5mm,
  left=1.5mm,right=1.5mm,top=1.1mm,bottom=1.1mm,
  before skip=5pt,after skip=6pt,
  listing options={basicstyle=\ttfamily\scriptsize,breaklines=true,columns=fullflexible,keepspaces=true,showstringspaces=false}
}
\newtcblisting{codebox}{
  listing only,
  enhanced,
  colback=CodeBack,
  colframe=CodeFrame,
  boxrule=0.55pt,
  arc=1.5mm,
  left=1.5mm,right=1.5mm,top=1.1mm,bottom=1.1mm,
  before skip=5pt,after skip=6pt,
  listing options={basicstyle=\ttfamily\scriptsize,breaklines=true,columns=fullflexible,keepspaces=true,showstringspaces=false}
}

\title{\bfseries F2A-ObjectNav: An Adaptive Agentic Architecture\\[0.15em]
for Object-Goal Navigation}

\author{
Seyed Ali Mirghasemi$^{a}$, Alireza Taheri$^{a,*}$\\[0.5em]
\small $^{a}$Social and Cognitive Robotics Laboratory, Mechanical Engineering Department,\\
\small Sharif University of Technology, Tehran, Iran\\[0.5em]
\small $^{*}$Corresponding author:
\href{mailto:artaheri@sharif.edu}{artaheri@sharif.edu};
Tel.: +98 21 6616 5531
}

\date{}

\newif\ifjournaltables
\journaltablestrue
\colorlet{TableHeaderText}{white}
\newcommand{\PlainTablePalette}[1]{%
  \colorlet{#1Header}{white}%
  \colorlet{#1Stripe}{white}%
  \colorlet{#1Rule}{black}%
}
\ifjournaltables
  \colorlet{TableHeaderText}{black}
  \forcsvlist{\PlainTablePalette}{StageI,Architecture,Evaluation,Ablation,Comparison,AppendixA,AppendixB,AppendixC,AppendixD}
\fi

\begin{document}

\maketitle

% ================================================================
% Abstract and Keywords
% ================================================================

\begin{abstract}
Object-goal navigation requires an embodied agent to explore an unfamiliar environment, identify a target object, navigate toward candidate instances, and reliably determine whether the observed object satisfies the mission goal. This work investigates this problem through a two-stage development process. In Stage I, a fixed modular framework is used to examine combinations of exploration, object detection, and target validation methods. The results reveal clear trade-offs among navigation performance, validation reliability, execution time, and model-related cost, with no single fixed configuration remaining preferable across all evaluated conditions. These observations motivate F2A-ObjectNav (Fixed-to-Adaptive Object-Goal Navigation), an adaptive agentic architecture in which a rule-based Supervisor coordinates specialized components for navigation and exploration, object detection, target validation, semantic guidance, and Mission Memory. The architecture uses event-driven recovery, strategy revision, candidate management, and conditional semantic guidance while retaining lightweight execution during normal operation. F2A-ObjectNav is evaluated over 25 episodes in each of AI2-THOR, RoboTHOR, and ProcTHOR, achieving success rate (SR) / success-weighted path length (SPL) values of 0.84/0.52, 0.40/0.22, and 0.44/0.37, respectively. In addition to task-level metrics, architecture-specific behavioral metrics are used to characterize semantic-guidance usage, strategy switching, recovery activation, memory-supported behavior, and decision stability. The results show different patterns of adaptive intervention across the three environments, with higher intervention frequencies observed in RoboTHOR and ProcTHOR than in AI2-THOR.
\end{abstract}

\vspace{0.5em}

\section{Introduction}
% ================================================================

Object-goal navigation (ObjectNav) requires an embodied agent to locate a target object in an unfamiliar environment without being given its position in advance \cite{batra2020objectnav}. Unlike point-goal navigation, where the destination is specified geometrically, ObjectNav requires the agent to infer where the target may be located, explore the environment, identify potential instances, navigate toward them, and determine whether an observed candidate satisfies the intended goal. Successful execution therefore depends on the coordinated operation of perception, exploration, navigation, and target verification over an extended sequence of decisions \cite{batra2020objectnav,chaplot2020gose}.

This coordination remains challenging because the information available to the agent is partial and changes throughout the mission. A target may initially be outside the field of view, object detections may be uncertain or incorrect, and visually similar objects may produce misleading candidates. Navigation can also become inefficient when the agent repeatedly visits previously explored regions, fails to reach a candidate, or continues using a search strategy that no longer produces useful progress. These difficulties make ObjectNav not only a perception or path-planning problem, but also a problem of deciding how different capabilities should be used as the mission evolves \cite{chaplot2020gose,ramakrishnan2022poni}.

Modular ObjectNav approaches address part of this challenge by separating functions such as semantic mapping, exploration, object detection, goal selection, and navigation into specialized components. Goal-Oriented Semantic Exploration, for example, combines semantic mapping with goal-directed exploration, while PONI explicitly separates the questions of where to search and how to navigate toward a selected location \cite{chaplot2020gose,ramakrishnan2022poni}. Such designs improve interpretability and allow individual modules to be developed or replaced independently. However, modularity alone does not provide mission-level adaptation. When the selection and interaction of these components are fixed before execution, the resulting pipeline may continue to apply the same strategy despite changes in mission conditions. A configuration that is effective during straightforward search may become less suitable after repeated exploration or ambiguous detections, whereas continuously invoking computationally expensive semantic reasoning may be unnecessary when conventional perception and navigation are sufficient.

Recent work has also demonstrated the value of pretrained semantic knowledge for target search. Methods such as L3MVN use large language models to provide object--scene priors for frontier selection, while ESC combines vision--language grounding with commonsense reasoning to support zero-shot exploration \cite{yu2023l3mvn,zhou2023esc}. These studies show that semantic knowledge can support search decisions when geometric exploration alone provides limited guidance. At the same time, such reasoning does not necessarily need to remain active throughout an entire mission. This motivates a complementary question: not only which semantic capabilities should be available, but also when they should be invoked and how they should interact with conventional perception and navigation components.

This question was first examined in Stage I of the present study through a fixed modular ObjectNav framework. Different combinations of exploration, detection, and target-validation methods were evaluated to characterize their trade-offs in navigation performance, validation reliability, task time, and model-related cost. The experiments showed that the relative advantages of the evaluated configurations changed with environmental complexity and mission conditions. Configurations incorporating semantic or multimodal reasoning often provided stronger validation reliability or navigation performance in more demanding conditions, whereas lighter configurations could operate with lower computational overhead. Importantly, no evaluated fixed configuration provided the most favorable balance across all considered criteria. These observations motivated investigating event-driven coordination as an alternative to committing to an unchanged combination of capabilities for the entire mission.

Motivated by these findings, this work introduces \textbf{F2A-ObjectNav}, an adaptive agentic architecture that extends modular ObjectNav with mission-level, event-driven coordination. The name F2A (Fixed-to-Adaptive) reflects the progression from the fixed modular configurations examined in Stage I to the adaptive architecture developed in Stage II. Rather than replacing conventional perception and navigation with a general-purpose reasoning model, the proposed architecture organizes specialized functions under a rule-based Supervisor that monitors mission progress and coordinates their use according to the current execution state. Mission Memory preserves relevant episode-level information, including rejected candidates and repeated or ineffective search behavior, while recovery and strategy revision provide mechanisms for responding to navigation or exploration failures. Semantic guidance is invoked selectively when the ongoing search does not provide sufficient progress, rather than operating continuously throughout the mission.

This design separates routine execution from adaptive intervention. Under normal conditions, navigation, exploration, and object detection proceed without requiring continuous high-level semantic reasoning. When events such as candidate rejection, navigation stagnation, repeated exploration, or insufficient search progress occur, the Supervisor can revise the current behavior, use previously stored mission information, activate recovery, or request semantic guidance. Detection is also separated from final target acceptance: detected objects are treated as candidates and are validated only after the agent reaches an appropriate observation point. The resulting architecture provides a structured mechanism for adapting the interaction among ObjectNav components while retaining their modularity, interpretability, and independent functional roles.

F2A-ObjectNav is evaluated in three complementary simulation environments: AI2-THOR \cite{kolve2017ai2thor}, RoboTHOR \cite{deitke2020robothor}, and ProcTHOR \cite{deitke2022procthor}. Standard ObjectNav performance is measured using Success Rate (SR) and Success-weighted Path Length (SPL), with SPL following the embodied-navigation evaluation formulation introduced by Anderson et al. \cite{anderson2018evaluation}. Additional system-level measures include task time, target-validation accuracy, and model-related cost. The evaluation further records architecture-specific behavioral indicators describing semantic-guidance usage, strategy switching, recovery activation, Mission Memory involvement, and supervisory decision stability. Together, these measures characterize both mission-level performance and the internal behavior of the adaptive coordination mechanism.

The main contributions of this work are:
\begin{enumerate}
    \item An empirical analysis of fixed modular ObjectNav configurations that characterizes the trade-offs among navigation performance, target-validation reliability, execution time, and model-related cost, and provides the experimental motivation for moving from fixed execution toward adaptive coordination.

    \item \textbf{F2A-ObjectNav}, an event-driven adaptive architecture for ObjectNav in which a rule-based Supervisor coordinates specialized components for navigation and exploration, object detection, target validation, semantic guidance, and episode-level Mission Memory.

    \item A set of mission-level adaptation mechanisms for responding to execution events through candidate management, memory-informed decisions, recovery, strategy revision, and conditional semantic guidance, without relying on a language model as the primary navigation controller.
    
    \item A multi-environment evaluation across AI2-THOR, RoboTHOR, and ProcTHOR that combines standard ObjectNav metrics, system-level performance measures, architecture-specific behavioral indicators, and an execution-trace mechanism analysis based on evaluation-wide statistics and recorded execution traces.
    
\end{enumerate}
% ================================================================
\section{Related Work}
% ================================================================

% ----------------------------------------------------------------
\subsection{Object-Goal Navigation Approaches}
% ----------------------------------------------------------------

Object-goal navigation has been studied through a range of end-to-end, modular, and increasingly open-vocabulary formulations \cite{sun2024survey}. These categories are not mutually exclusive; rather, they describe complementary choices in how perception, navigation, semantic reasoning, and policy learning are organized. End-to-end methods typically learn navigation behavior directly from observations and target information by integrating perception, representation learning, and action selection within a single policy. Although such approaches provide a compact formulation, their behavior is closely tied to the learned representation and training distribution, which can complicate generalization and failure analysis \cite{sun2024survey}.

Modular approaches instead decompose the navigation problem into explicit components such as mapping, exploration, goal selection, path planning, and perception. Semantic mapping provides a broader basis for linking geometric representations to knowledge representation and task planning \cite{kostavelis2015semantic}. Goal-Oriented Semantic Exploration constructs an episodic semantic map and predicts long-term exploration goals according to the target category \cite{chaplot2020gose}. PONI follows a related modular formulation but learns potential functions over semantic maps to estimate promising and reachable search regions \cite{ramakrishnan2022poni}. These methods demonstrate the advantages of explicitly separating navigation capabilities, particularly for long-horizon exploration and interpretable decision making. However, the interaction among their major components is generally specified by the system design and remains largely fixed during execution.

Other work has introduced specialization within the navigation policy itself. DAT distinguishes between searching for an unseen target and navigating toward a previously observed one, adapting its reasoning strategy to these different stages \cite{dang2023dat}. MT further separates multiple navigation abilities and coordinates them through a meta-ability decoupling framework \cite{dang2023mt}. These studies indicate that different navigation situations can benefit from different behaviors and motivate mechanisms that revise the active behavior as the mission evolves.

% ----------------------------------------------------------------
\subsection{Zero-Shot and Semantic-Guided Navigation}
% ----------------------------------------------------------------

Recent research has increasingly focused on zero-shot and open-vocabulary ObjectNav, where pretrained visual and language representations reduce dependence on task-specific navigation training. ZSON transfers multimodal representations learned through image-goal navigation to ObjectNav, enabling navigation toward targets specified through language without requiring ObjectNav demonstrations \cite{majumdar2022zson}. CoW takes a more directly training-free approach by combining open-vocabulary visual localization with conventional exploration mechanisms \cite{gadre2023cows}.

A related direction uses semantic knowledge to determine where the agent should search. Earlier active visual search systems used topological spatial relations to guide object search \cite{sjoo2012topological}. L3MVN incorporates object--scene relationships derived from a large language model into frontier selection \cite{yu2023l3mvn}, while ESC combines open-vocabulary visual grounding with commonsense reasoning to prioritize semantically plausible search regions \cite{zhou2023esc}. VLFM similarly augments geometric frontier exploration with a vision--language value map that estimates the semantic relevance of unexplored regions \cite{yokoyama2024vlfm}. More recent work has continued this direction. Sun et al.\ incorporate LLM-derived room--object relationships into a modular map-based ObjectNav framework \cite{sun2025roomobject}, while Xie et al.\ use LLM-based object--room knowledge to guide diffusion-based reasoning over unexplored semantic maps \cite{xie2026dar}. STG-Nav combines a structured scene representation with LLM-based semantic reasoning for zero-shot target search \cite{lin2026stgnav}.

These methods demonstrate that pretrained semantic knowledge can substantially improve target-location reasoning and search-region selection. Their primary focus, however, is generally on improving semantic exploration, long-term goal prediction, or target localization. The present work addresses a complementary problem: deciding when semantic reasoning should be invoked and how it should interact with detection, navigation, validation, recovery, and mission-level memory during execution.

Large language models have also been used more directly for high-level navigation reasoning. SayNav incrementally constructs a semantic scene representation and uses an LLM to generate and revise high-level navigation plans as new observations become available \cite{rajvanshi2024saynav}. These approaches demonstrate the usefulness of language-based reasoning for planning and semantic search in previously unseen environments, while also motivating architectures in which the role of high-level reasoning is explicitly separated from low-level navigation control.

% ----------------------------------------------------------------
\subsection{Memory, Adaptation, and Agent-Based Coordination}
% ----------------------------------------------------------------

Long-horizon navigation requires an agent to retain information from previous observations and decisions. DAT uses target-oriented memory to support navigation after a target has been observed \cite{dang2023dat}, while Memory-MambaNav introduces a spatial--temporal state-space representation for efficiently processing accumulated navigation history \cite{sun2025memorymamba}. These approaches illustrate the importance of temporal information and memory for preserving useful context during extended ObjectNav episodes.

Adaptation has also been introduced at different levels of the navigation process. DAT changes behavior between target search and target-directed navigation \cite{dang2023dat}, while MT coordinates several specialized navigation abilities \cite{dang2023mt}. More recently, CogNav models ObjectNav as a sequence of cognitive states ranging from broad exploration to candidate verification and target confirmation. A large language model determines transitions among these states using information from a dynamically updated cognitive map \cite{cao2025cognav}. This formulation provides an explicit mechanism for changing navigation behavior according to the evolving state of the mission.

Dynamic failure handling is also central to TriHelper, which introduces specialized assistance for collision, exploration, and detection, including a false-target map for retaining rejected target information \cite{zhang2024trihelper}. Thus, failure-triggered assistance and rejection memory are established design ideas rather than individually novel elements of F2A-ObjectNav. The focus here is their organization within an explicit supervisory workflow, with separate detection and goal-aware validation interfaces and episode-level execution records.

F2A-ObjectNav differs from this line of work in the role assigned to high-level reasoning. In CogNav, an LLM is directly involved in determining transitions between cognitive states. In F2A-ObjectNav, mission-level coordination is instead handled by an explicit rule-based Supervisor. Language and multimodal models are assigned specialized functions---such as semantic guidance and target validation---and are invoked conditionally rather than serving as the primary supervisory controller. The architecture additionally treats candidate rejection, recovery, strategy revision, and episode-level Mission Memory as explicit events within the supervisory process.

More broadly, agent-based systems provide a mechanism for organizing specialized reasoning components. ReAct introduced an interaction pattern in which reasoning and actions are interleaved so that new observations can modify subsequent decisions \cite{yao2023react}. In robot control, LLM-Controller uses language models to support dynamic adaptation of conventional controllers \cite{zahedifar2025llmcontroller}. LLM-Agent-Controller organizes multiple specialized agents under supervisory coordination for control-engineering tasks \cite{zahedifar2025llmagentcontroller}. Although these systems address different tasks, they illustrate the broader value of role specialization, feedback, and supervisory coordination in multi-step decision-making architectures.

% ----------------------------------------------------------------
\subsection{Positioning of the Present Work}
% ----------------------------------------------------------------

Existing ObjectNav research has established effective mechanisms for semantic mapping, exploration, zero-shot target recognition, episodic memory, language-guided planning, and adaptive navigation behavior. Several approaches also coordinate multiple capabilities or respond explicitly to navigation failures. The distinction pursued here is therefore the particular supervisory organization and its documented execution behavior, rather than adaptation or memory in isolation.

The focus of the present work is the coordination problem itself. F2A-ObjectNav extends a fixed modular ObjectNav framework with an event-driven supervisory layer that coordinates navigation and exploration, object detection, target validation, Mission Memory, recovery, strategy revision, and conditional semantic guidance. The proposed architecture therefore does not aim to replace established perception, exploration, or navigation techniques. Instead, it provides a structured mechanism for changing how these capabilities are used in response to mission-level events.

A central design distinction is that the Supervisor remains rule-based rather than delegating mission-level control to a large language model. Language and multimodal models are used only for specialized tasks where semantic interpretation is required. This separation enables conventional perception and navigation modules to handle routine execution while semantic reasoning, recovery, memory-based intervention, or strategy revision is activated only when the current mission state requires it. The resulting architecture combines the interpretability and replaceability of modular navigation with event-driven adaptation motivated directly by the limitations observed in the fixed configurations studied in Stage I.

% ================================================================
\section{Problem Formulation}
% ================================================================

In object-goal navigation, an embodied agent is given a goal $G$, specified as an object category or a natural-language description, and must locate a corresponding object in an unfamiliar environment \cite{batra2020objectnav}. Because the complete state of the environment is not known in advance, the agent must act on partial observations collected during navigation and progressively gather information about the scene.

Let $o_t \in \mathcal{O}$ denote the observation available at time step $t$, and let $a_t \in \mathcal{A}$ denote the corresponding action. Depending on the evaluation environment, an observation may include RGB images, depth information, pose information, or other available sensory inputs, while the action space may include translational motion, rotation, viewpoint adjustment, and a stop action. The information available to the agent at time $t$ can be represented by the interaction history
\begin{equation}
h_t = (o_1,a_1,o_2,a_2,\ldots,o_t).
\end{equation}

A decision mechanism selects the next action from the current mission history and the specified goal:
\begin{equation}
a_t = \pi(G,h_t), \qquad a_t \in \mathcal{A}.
\end{equation}

The mission terminates when the system accepts a candidate as the final target and stops, or when the execution-step budget is exhausted. Termination does not by itself imply success. In the Stage II evaluation, success requires the correct target to be visible and within 1~m of the agent's camera at stopping, with target correctness assessed by human inspection. Incorrect target acceptance and budget exhaustion without successful completion are scored as failures.

The objective is therefore to complete the ObjectNav mission efficiently and reliably under partial observability. In this work, system performance is characterized using task-success and path-efficiency measures together with execution time, target-validation accuracy, and model-related cost. The detailed definitions of these measures are provided in the experimental evaluation section.

% ================================================================
\section{Methodology}
% ================================================================

\begin{figure}[H]
    \centering
    \IfFileExists{figures/fig1.pdf}{
        \includegraphics[width=0.92\linewidth,height=0.80\textheight,keepaspectratio]{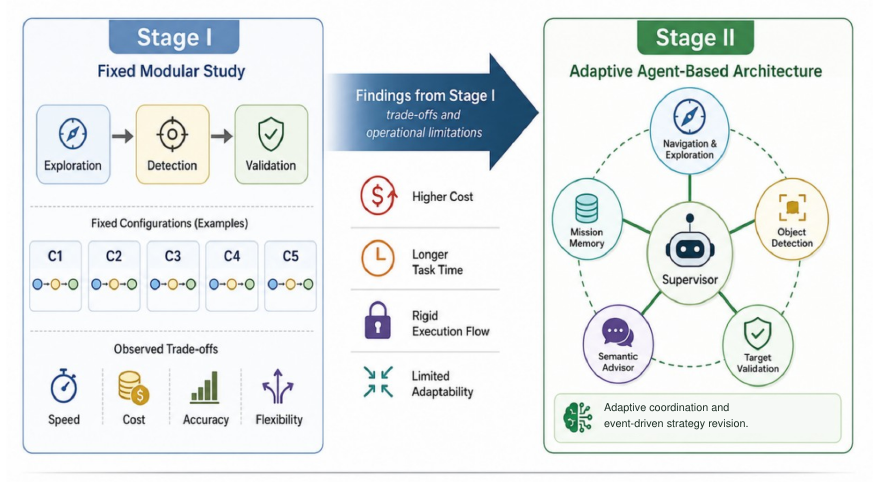}
    }{
        \fbox{\parbox[c][4.0cm][c]{0.88\linewidth}{
            \centering
            \textbf{Figure placeholder}\\[4pt]
            Research progression from the fixed modular study to the adaptive architecture\\[4pt]
            \texttt{figures/fig1.pdf}
        }}
    }
    \caption{Research progression from the fixed modular study to the proposed adaptive agentic architecture. (AI-generated schematic using ChatGPT.)}
    \label{fig:research_progression}
\end{figure}

This study follows a two-stage methodological progression. Stage I examines whether a fixed combination of exploration, object detection, and target validation modules can provide a consistent balance among navigation performance, execution time, validation reliability, and model-related cost. The findings of this analysis then motivate Stage II, in which the fixed modular structure is extended with an adaptive coordination architecture. The fixed-configuration results provide a design motivation for adaptive coordination; they do not establish a performance advantage for the subsequent architecture. The overall progression of the study is summarized in Fig.~\ref{fig:research_progression}.

% ----------------------------------------------------------------
\subsection{Stage I: Fixed Modular Study}
% ----------------------------------------------------------------

Stage I builds on a preliminary fixed modular ObjectNav study presented at ICRoM 2025 \cite{mirghasemi2025fixed}. The earlier study focused on comparing representative combinations of exploration, object detection, and target-validation modules under different environmental conditions. In the present work, these findings are used as the empirical basis for the development of the adaptive architecture introduced in Stage II. The journal study substantially extends the earlier fixed-configuration analysis by introducing mission-level adaptive coordination, a rule-based Supervisor, Mission Memory, recovery and strategy-revision mechanisms, conditional semantic guidance, and evaluation across AI2-THOR, RoboTHOR, and ProcTHOR.

% ----------------------------------------------------------------
\subsubsection{Fixed Pipeline and Navigation Backbone}
% ----------------------------------------------------------------

\begin{figure}[H]
    \centering
    \IfFileExists{figures/fig2.pdf}{
        \includegraphics[width=0.18\linewidth,keepaspectratio]{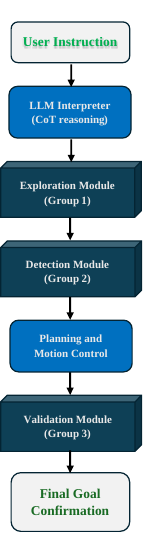}
    }{
        \fbox{\parbox[c][4.2cm][c]{0.88\linewidth}{
            \centering
            \textbf{Figure placeholder}\\[4pt]
            Architecture of the fixed modular framework (Stage I)\\[4pt]
            \texttt{figures/fig2.pdf}
        }}
    }
    \caption{Architecture of the fixed modular framework used in Stage I.}
    \label{fig:stage1_architecture}
\end{figure}

The Stage I framework follows a sequential pipeline consisting of goal interpretation, environment exploration, object detection, candidate navigation, target validation, and mission termination, as illustrated in Fig.~\ref{fig:stage1_architecture}. At the beginning of an episode, the user provides an object category or a natural-language target description. Goal interpretation extracts a concise target label for detection while retaining the complete instruction for final validation.

During exploration, visual observations are processed by the active detection module. When a target candidate is identified, its estimated location is passed to the navigation component, which moves the agent to a closer observation point. The validation module then determines whether the candidate satisfies the original goal. A positive validation terminates the mission; otherwise, exploration resumes using the same active configuration.

Mapping, localization, path planning, obstacle avoidance, and motion control are kept unchanged across all configurations. This common navigation backbone allows the comparison to focus on the higher-level exploration, detection, and validation choices rather than on differences in low-level navigation.

% ----------------------------------------------------------------
\subsubsection{Variable Modules and Fixed Configurations}
% ----------------------------------------------------------------

Three higher-level modules are varied in Stage I: exploration, object detection, and target validation. Each module contains alternatives that differ in semantic capability, computational demand, and dependence on predefined object classes.

The exploration alternatives include random exploration, frontier-based exploration \cite{yamauchi1997frontier}, BLIP-2-guided exploration \cite{li2023blip2}, and GPT-4o-guided exploration \cite{openai2024gpt4o}. Random exploration provides a low-cost reference without explicit map- or semantic-based guidance. Frontier-based exploration systematically directs the agent toward boundaries between explored and unexplored regions. BLIP-2 and GPT-4o provide different forms of semantic guidance based on the visual scene and target description.

For object detection, the framework uses YOLO11 as a closed-set detector \cite{khanam2024yolov11}, YOLO-World for open-vocabulary detection \cite{cheng2024yoloworld}, and BLIP-2 as a vision--language matching component \cite{li2023blip2}. A detection is treated only as a target candidate and does not by itself indicate mission success.

Target validation is performed using YOLO-World, BLIP-2, or GPT-4o. These alternatives provide different mechanisms for assessing whether an observed candidate is consistent with the specified target, while also introducing different computational and model-related costs.

The study does not evaluate the full Cartesian product of all module alternatives. Instead, five automatic configurations are selected to represent distinct combinations of lightweight perception, open-vocabulary detection, vision--language matching, and multimodal reasoning. The purpose of this comparison is to examine system-level trade-offs among representative fixed pipelines rather than to isolate the independent causal contribution of each individual module. A manually controlled condition is included separately as a human-reference condition. Table~\ref{tab:stage1_configs} summarizes the evaluated configurations.

\begin{table}[!ht]
    \centering
    \caption{Composition of the fixed modular configurations evaluated in Stage I.}
    \label{tab:stage1_configs}
    \footnotesize
    \StageITableStyle
    \begin{tabularx}{\linewidth}{C{0.12\linewidth}ZZZ}
        \toprule
        \rowcolor{StageIHeader}
        \textcolor{TableHeaderText}{\textbf{Config.}} & \textcolor{TableHeaderText}{\textbf{Exploration}} & \textcolor{TableHeaderText}{\textbf{Detection}} & \textcolor{TableHeaderText}{\textbf{Validation}} \\
        \midrule
        C1 & Random & YOLO11 & YOLO-World \\
        C2 & Frontier-based & YOLO-World & GPT-4o \\
        C3 & Frontier-based & BLIP-2 & YOLO-World \\
        C4 & BLIP-2-guided & YOLO-World & GPT-4o \\
        C5 & GPT-4o-guided & YOLO-World & BLIP-2 \\
        C6 & \multicolumn{3}{c}{Manual control (human-reference condition)} \\
        \bottomrule
    \end{tabularx}
\end{table}

The active configuration is selected before each episode and remains unchanged until success or termination. In particular, rejecting a candidate causes the system to resume exploration but does not replace the active exploration, detection, or validation method. This fixed execution behavior is the central property examined in Stage I and provides the basis for the adaptive coordination introduced in Stage II.

% ----------------------------------------------------------------
\subsubsection{Evaluation Protocol}
% ----------------------------------------------------------------

The fixed modular configurations were evaluated in Gazebo using two levels of environmental complexity. The simple layout consisted of a single room, whereas the complex layout contained three connected rooms with the same object set distributed across multiple areas. Fifteen runs were conducted for each of the five automatic configurations in each layout, resulting in 150 automatic runs in total.

The manually controlled C6 condition was evaluated separately as a human-reference condition. Its purpose was to provide an approximate reference for task completion under manual control rather than to serve as an additional automatic modular configuration. Accordingly, C6 is reported separately from the five automatic configurations when interpreting the Stage I results.

\begin{figure}[H]
    \centering
    \IfFileExists{figures/fig3.pdf}{
        \includegraphics[width=0.96\linewidth,height=0.80\textheight,keepaspectratio]{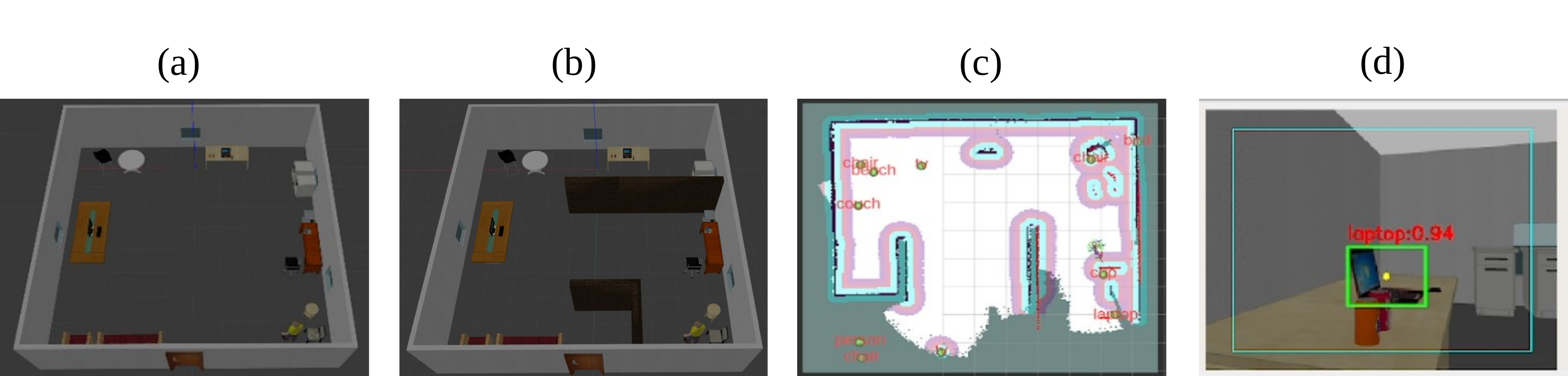}
    }{
        \fbox{\parbox[c][4.8cm][c]{0.92\linewidth}{
            \centering
            \textbf{Figure placeholder}\\[4pt]
            Stage I simulation environments and representative outputs:\\
            (a) simple layout, (b) complex layout,\\
            (c) occupancy map, and (d) detected target candidate\\[4pt]
            \texttt{figures/fig3.pdf}
        }}
    }
    \caption{Stage I simulation environments and representative outputs: (a) simple single-room layout, (b) complex three-room layout, (c) occupancy map produced during frontier-based exploration, and (d) first-person observation showing a detected target candidate.}
    \label{fig:stage1_environment}
\end{figure}

Figure~\ref{fig:stage1_environment} shows the two simulation layouts together with representative mapping and detection outputs.

Stage I evaluates the fixed configurations from several complementary perspectives. Success Rate (SR) and Success-weighted Path Length (SPL) measure mission completion and path efficiency, respectively. The Environmental Adaptability Score (EAS) summarizes the relative retention of SR and SPL when moving from the simple layout to the more complex layout. Task Time (TT) measures episode duration, Validation Accuracy (VA) measures the correctness of candidate-validation decisions, and API Cost (AC) captures the model-related inference cost associated with externally accessed models.

The metrics used in Stage I are summarized in Table~\ref{tab:stage1_metrics}. Their mathematical definitions and calculation procedures are provided in Appendix~\ref{app:metric_definitions}. The corresponding quantitative results are presented in the following subsection and are subsequently used to identify the limitations of fixed execution and motivate the adaptive architecture introduced in Stage II.

\begin{table}[!ht]
    \centering
    \caption{Metrics used to evaluate the Stage I configurations.}
    \label{tab:stage1_metrics}
    \footnotesize
    \StageITableStyle
    \begin{tabularx}{\linewidth}{C{0.38\linewidth}C{0.10\linewidth}Z}
        \toprule
        \rowcolor{StageIHeader}
        \textcolor{TableHeaderText}{\textbf{Metric}} & \textcolor{TableHeaderText}{\textbf{Symbol}} & \textcolor{TableHeaderText}{\textbf{Evaluation purpose}} \\
        \midrule
        Success Rate
        & SR
        & Proportion of successfully completed missions \\

        Success-weighted Path Length
        & SPL
        & Mission success weighted by path efficiency \\

        Environmental Adaptability Score
        & EAS
        & \RaggedRight Relative retention of SR and SPL from the simple to the complex layout \\

        Task Time
        & TT
        & Average duration from mission initialization to termination \\

        Validation Accuracy
        & VA
        & Correctness of candidate-validation decisions \\

        API Cost
        & AC
        & Average model-related inference cost per episode \\
        \bottomrule
    \end{tabularx}
\end{table}

% ----------------------------------------------------------------
\subsubsection{Stage I Results}
% ----------------------------------------------------------------

The Stage I results were examined to determine whether any fixed configuration provided a consistently favorable trade-off among navigation effectiveness, validation reliability, execution time, and model-related cost. Tables~\ref{tab:stage1_navigation_results} and \ref{tab:stage1_efficiency_results} report the results of the preliminary Stage I study \cite{mirghasemi2025fixed}.

Table~\ref{tab:stage1_navigation_results} reports Success Rate (SR), Success-weighted Path Length (SPL), and Environmental Adaptability Score (EAS) for the simple and complex layouts. EAS is calculated from the relative retention of SR and SPL when moving from the simple layout to the complex layout, as defined in Appendix~\ref{app:metric_definitions}.

Among the automatic configurations, C4 achieved the strongest navigation results overall. It reached an SR of 1.00 in the simple layout and 0.87 in the complex layout, together with SPL values of 0.76 and 0.54, respectively. C4 also obtained the highest EAS among the automatic configurations, with a value of 0.790, indicating the strongest relative retention of navigation performance as environmental complexity increased. C5 also performed strongly, with an EAS of 0.759, although its complex-layout SR and SPL were slightly lower than those of C4.

C2 and C3 showed similar levels of performance retention, with EAS values of 0.593 and 0.595, respectively, despite differing in their detection and validation components. C1 exhibited the largest degradation between the two layouts, with its SR decreasing from 0.73 to 0.27 and its SPL from 0.51 to 0.09, resulting in an EAS of 0.273. The manually controlled C6 condition achieved the highest absolute navigation values, but it is included only as a human-reference condition and is not treated as an automatic configuration.

\begin{table}[!ht]
    \centering
    \caption{Navigation effectiveness of the fixed configurations in the simple and complex layouts.}
    \label{tab:stage1_navigation_results}
    \footnotesize
    \StageITableStyle
    \begin{tabularx}{\linewidth}{C{0.12\linewidth}*{5}{Z}}
        \toprule
        \rowcolor{StageIHeader}
        & \multicolumn{2}{c}{\textcolor{TableHeaderText}{\textbf{SR}}} & \multicolumn{2}{c}{\textcolor{TableHeaderText}{\textbf{SPL}}} & \\
        \noalign{\kern-0.2pt} % Overlap adjacent header fills to suppress PDF hairlines.
        \rowcolor{StageIHeader}
        \multirow{-2}{*}{\textcolor{TableHeaderText}{\textbf{Config.}}} & \textcolor{TableHeaderText}{\textbf{Simple}} & \textcolor{TableHeaderText}{\textbf{Complex}} & \textcolor{TableHeaderText}{\textbf{Simple}} & \textcolor{TableHeaderText}{\textbf{Complex}} & \multirow{-2}{*}{\textcolor{TableHeaderText}{\textbf{EAS}}} \\
        \midrule
        C1 & 0.73 & 0.27 & 0.51 & 0.09 & 0.273 \\
        C2 & 0.87 & 0.67 & 0.60 & 0.25 & 0.593 \\
        C3 & 0.87 & 0.60 & 0.58 & 0.29 & 0.595 \\
        C4 & \textbf{1.00} & \textbf{0.87} & \textbf{0.76} & \textbf{0.54} & \textbf{0.790} \\
        C5 & \textbf{1.00} & 0.80 & 0.71 & 0.51 & 0.759 \\
        \midrule
        C6 & 1.00 & 1.00 & 0.86 & 0.67 & 0.890 \\
        \bottomrule
    \end{tabularx}
\end{table}

The efficiency, validation, and model-related cost results in Table~\ref{tab:stage1_efficiency_results} reveal a different ordering. C4 achieved the shortest task time among the automatic configurations in the simple layout, completing missions in an average of 74~s, whereas C2 obtained the shortest task time in the complex layout at 281~s. C2 and C4 also achieved the highest Validation Accuracy among the automatic configurations, both reaching 0.91.

C1 and C3 incurred no external API cost, but their Validation Accuracy was lower at 0.73. Their navigation behavior, however, differed substantially: C1 showed a pronounced reduction in both SR and SPL in the complex layout, whereas C3 retained a larger proportion of its simple-layout performance. C5 achieved strong navigation performance but introduced the highest API-related cost and the longest task time in the complex layout, reaching 383~s and an average API cost of \$0.0079 per episode.

\begin{table}[!ht]
    \centering
    \caption{Efficiency, model-related cost, and validation results of the Stage I configurations.}
    \label{tab:stage1_efficiency_results}
    \footnotesize
    \StageITableStyle
    \begin{tabularx}{\linewidth}{C{0.12\linewidth}*{5}{Z}}
        \toprule
        \rowcolor{StageIHeader}
        & \multicolumn{2}{c}{\textcolor{TableHeaderText}{\textbf{TT (s)}}} & \multicolumn{2}{c}{\textcolor{TableHeaderText}{\textbf{API Cost (\$)}}} & \\
        \noalign{\kern-0.2pt} % Overlap adjacent header fills to suppress PDF hairlines.
        \rowcolor{StageIHeader}
        \multirow{-2}{*}{\textcolor{TableHeaderText}{\textbf{Config.}}} & \textcolor{TableHeaderText}{\textbf{Simple}} & \textcolor{TableHeaderText}{\textbf{Complex}} & \textcolor{TableHeaderText}{\textbf{Simple}} & \textcolor{TableHeaderText}{\textbf{Complex}} & \multirow{-2}{*}{\textcolor{TableHeaderText}{\textbf{VA}}} \\
        \midrule
        C1 & 211 & 304 & \textbf{0.0000} & \textbf{0.0000} & 0.73 \\
        C2 & 105 & \textbf{281} & 0.0015 & 0.0015 & \textbf{0.91} \\
        C3 & 93 & 311 & \textbf{0.0000} & \textbf{0.0000} & 0.73 \\
        C4 & \textbf{74} & 367 & 0.0015 & 0.0015 & \textbf{0.91} \\
        C5 & 79 & 383 & 0.0031 & 0.0079 & 0.84 \\
        \midrule
        C6 & 19 & 31 & 0.0000 & 0.0000 & 1.00 \\
        \bottomrule
    \end{tabularx}
\end{table}

Taken together, the Stage I results reveal a trade-off among navigation effectiveness, validation reliability, execution time, and model-related cost rather than a universally preferable fixed pipeline. C4 provided the strongest overall navigation performance and the highest performance retention among the automatic configurations, while C2 achieved lower execution time in the complex environment with the same Validation Accuracy. C3 avoided external API cost and retained a similar proportion of navigation performance to C2, but with lower Validation Accuracy. C5 provided strong navigation performance at the expense of greater execution time and model-related cost.

Because several modules change simultaneously across the evaluated configurations, these results should not be interpreted as isolating the independent causal contribution of any individual exploration, detection, or validation component. Instead, Stage I demonstrates that representative fixed pipelines occupy different positions in a performance--efficiency--cost trade space and that the preferred configuration depends on the evaluation criterion and environmental condition.

\begin{figure}[H]
    \centering
    \IfFileExists{figures/fig4.pdf}{
        \includegraphics[width=0.90\linewidth,keepaspectratio]{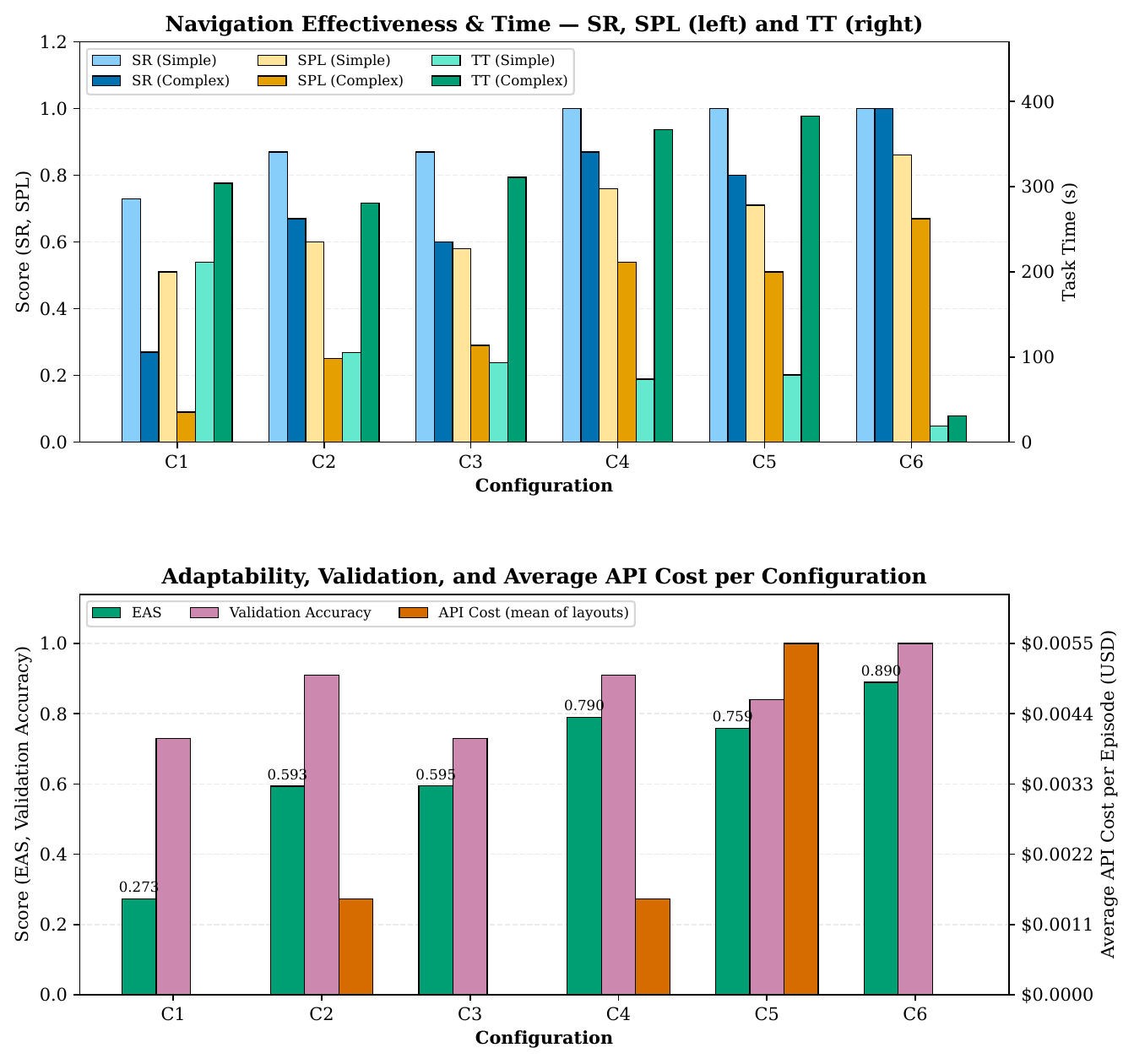}
    }{
        \fbox{\parbox[c][5.0cm][c]{0.92\linewidth}{
            \centering
            \textbf{Figure placeholder}\\[4pt]
            Comparative Stage I results\\
            (Top) SR, SPL, and Task Time\\
            (Bottom) EAS, Validation Accuracy, and average API Cost\\[4pt]
            \texttt{figures/fig4.pdf}
        }}
    }
    \caption{Comparative results of Stage I. Top: SR and SPL for each configuration, with Task Time shown on a secondary axis. Bottom: EAS and Validation Accuracy (left axis), with API Cost averaged equally across the two layouts on the right axis. EAS is calculated from Table~\ref{tab:stage1_navigation_results}.}
    \label{fig:stage1_results}
\end{figure}

The principal numerical trends are summarized in Fig.~\ref{fig:stage1_results}. The manually controlled C6 condition is shown only as a human-reference condition and is not treated as an automatic configuration.

% ----------------------------------------------------------------
\subsection{From Fixed Configurations to Adaptive Coordination}
% ----------------------------------------------------------------

The Stage I trade-offs motivate investigating whether an exploration--detection--validation pipeline can respond to changing mission conditions instead of retaining one configuration throughout an episode. This is a design hypothesis rather than a causal conclusion from the fixed-configuration comparison. A configuration that is effective under one set of conditions may become less suitable when the search becomes prolonged, a candidate is rejected, navigation progress deteriorates, or the active exploration strategy repeatedly visits unproductive regions.

Four design limitations are particularly relevant to the development of Stage II. First, the fixed pipeline does not distinguish among different types of mission difficulty when deciding how execution should continue. Candidate rejection, repeated exploration, navigation stagnation, and failure to approach a candidate all return control to essentially the same predefined execution process. Second, information accumulated during an episode---such as rejected candidates, repeatedly visited regions, unsuccessful navigation attempts, or ineffective recovery actions---is not maintained in a structured mission-level memory that can directly influence later decisions. Third, semantic reasoning is tied to the selected configuration rather than invoked according to the current need. Consequently, semantically richer modules may remain active even when conventional perception and navigation are sufficient, whereas a lightweight configuration cannot request additional semantic support when ambiguity arises. Fourth, the fixed framework lacks a supervisory mechanism for selecting different responses to different execution events.

These observations motivate a different coordination principle for Stage II. Instead of searching for a single fixed pipeline, the system should monitor mission progress, retain relevant episode-level information, distinguish among execution events, and revise its behavior when the current strategy becomes ineffective. Semantic reasoning should likewise be treated as a conditional resource rather than a continuously active controller.

Table~\ref{tab:stage1_to_stage2} summarizes how the main limitations identified in Stage I motivate the design choices introduced in Stage II.

\begin{table}[!ht]
    \centering
    \caption{Design transition from the limitations of the fixed framework to the adaptive mechanisms introduced in Stage II.}
    \label{tab:stage1_to_stage2}
    \footnotesize
    \StageITableStyle
    \begin{tabularx}{\linewidth}{C{0.44\linewidth}Z}
        \toprule
        \rowcolor{StageIHeader}
        \textcolor{TableHeaderText}{\textbf{Stage I limitation}} & \textcolor{TableHeaderText}{\textbf{Stage II design response}} \\
        \midrule

        One configuration remains active throughout the mission &
        Rule-based supervisory coordination and conditional strategy revision \\

        \RaggedRight Different failure events return to the same predefined execution process &
        Event-specific recovery and mission-level decision logic \\

        \RaggedRight Rejected candidates and unsuccessful search history are not systematically reused &
        Mission Memory and memory-informed candidate management \\

        \RaggedRight Semantic reasoning is determined by the selected configuration rather than current need &
        Conditional activation of the Semantic Advisor \\

        \bottomrule
    \end{tabularx}
\end{table}

Accordingly, Stage II introduces F2A-ObjectNav, an event-driven adaptive architecture in which a rule-based Supervisor coordinates specialized navigation, perception, validation, memory, recovery, and semantic-reasoning components according to the current mission state. The architecture retains the modularity of the Stage I framework while adding a mission-level coordination layer that determines when additional intervention is required.

% ================================================================
\section{F2A-ObjectNav: Proposed Adaptive Architecture}
% ================================================================

% ----------------------------------------------------------------
\subsection{System Overview}
% ----------------------------------------------------------------

F2A-ObjectNav is designed as a modular, event-driven architecture for ObjectNav in which routine perception and navigation remain separated from mission-level adaptive intervention. The architecture contains six main elements:

\begin{enumerate}
    \item Supervisor Agent,
    \item Navigation and Exploration Agent,
    \item Object Detection Agent,
    \item Target Validation Agent,
    \item Semantic Advisor Agent, and
    \item Mission Memory.
\end{enumerate}

Here, an agent denotes a role-specific software component with an explicit input--output interface, not an independently learned policy or a separate physical robot. Adaptation refers to event-driven changes in execution behavior within an episode; it does not imply online parameter learning.

The Supervisor Agent coordinates mission execution through rules rather than LLM-generated control decisions. At initialization, an LLM-based Target Interpreter extracts the object category from the user request and prepares a validation description that preserves the requested attributes and details. This input-processing role is distinct from the Semantic Advisor, which provides conditional guidance during search. Mission Memory is then initialized, and the Supervisor activates the default exploration and detection process.

During normal execution, the Navigation and Exploration Agent searches the environment while the Object Detection Agent analyzes incoming observations. A detection is treated as a candidate rather than as evidence of mission completion. When a candidate is selected, the Navigation and Exploration Agent approaches an appropriate observation point, after which the Target Validation Agent evaluates whether the candidate satisfies the complete target specification.

Adaptive intervention is introduced only when the current execution state requires it. Events such as candidate rejection, repeated exploration, navigation stagnation, unsuccessful candidate approach, or prolonged lack of progress can cause the Supervisor to update Mission Memory, activate recovery, revise the exploration strategy, or request assistance from the Semantic Advisor. The Semantic Advisor provides high-level contextual guidance but does not issue low-level navigation commands.

\begin{figure}[H]
    \centering
    \IfFileExists{figures/fig5.pdf}{
        \includegraphics[width=0.7\linewidth,keepaspectratio]{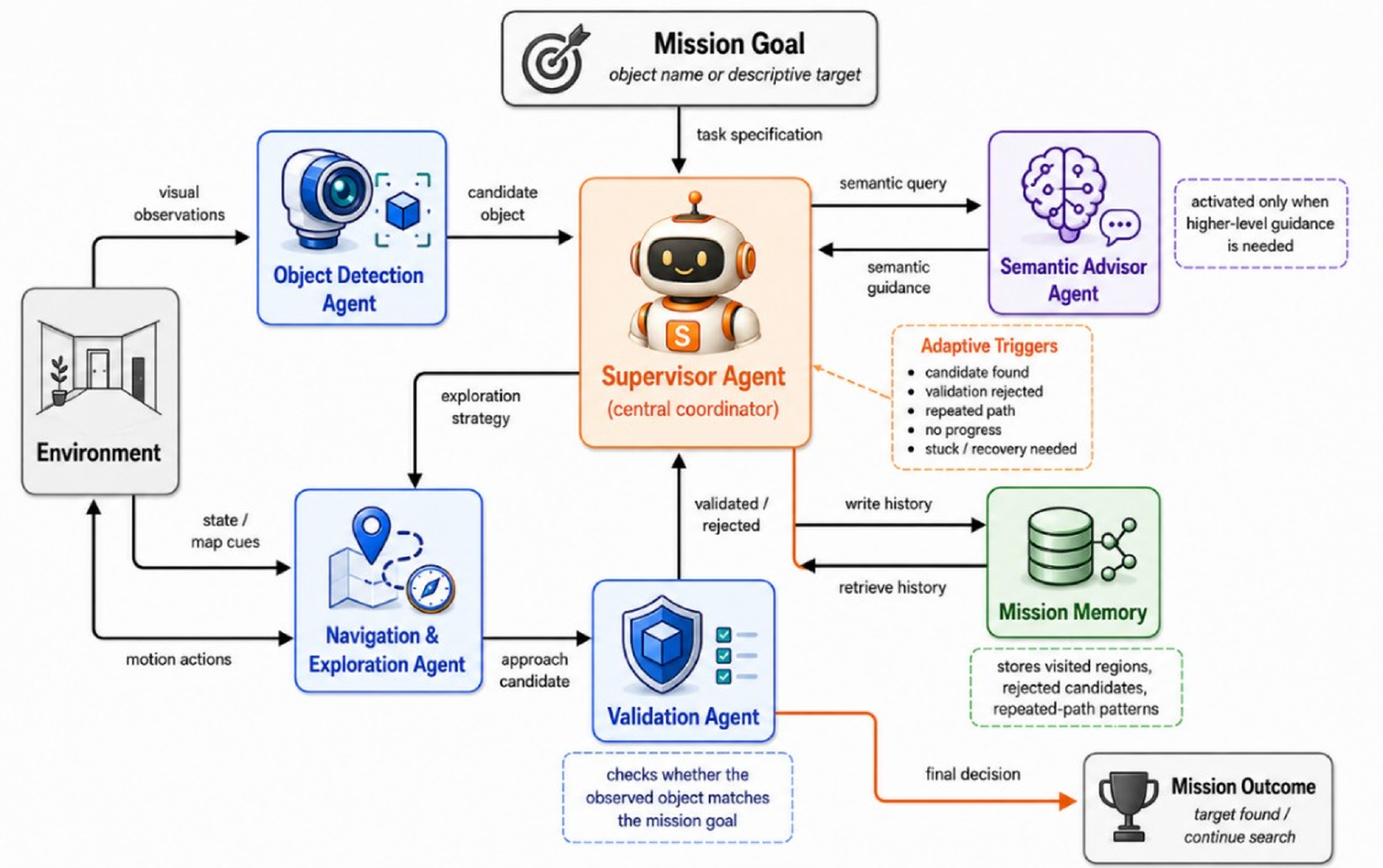}
    }{
        \fbox{\parbox[c][5.2cm][c]{0.92\linewidth}{
            \centering
            \textbf{Figure placeholder}\\[4pt]
            Overall architecture of F2A-ObjectNav\\[4pt]
            Rule-based Supervisor coordinating navigation and exploration,\\
            object detection, target validation, Semantic Advisor, and Mission Memory\\[4pt]
            \texttt{figures/fig5.pdf}
        }}
    }
    \caption{Overall architecture of F2A-ObjectNav. A rule-based Supervisor coordinates specialized ObjectNav components and conditionally activates memory, recovery, strategy revision, and semantic guidance according to mission-level events. (AI-generated schematic using ChatGPT.)}
    \label{fig:f2a_architecture}
\end{figure}

The overall architecture is illustrated in Fig.~\ref{fig:f2a_architecture}.

Table~\ref{tab:f2a_components} summarizes the functional organization of the architecture.

\begin{table}[!ht]
    \centering
    \caption{Functional organization of the main F2A-ObjectNav components.}
    \label{tab:f2a_components}
    \footnotesize
    \ArchitectureTableStyle
    \begin{tabularx}{\linewidth}{C{0.21\linewidth}C{0.22\linewidth}C{0.22\linewidth}Z}
        \toprule
        \rowcolor{ArchitectureHeader}
        \textcolor{TableHeaderText}{\textbf{Component}} &
        \textcolor{TableHeaderText}{\textbf{Main input}} &
        \textcolor{TableHeaderText}{\textbf{Main output}} &
        \textcolor{TableHeaderText}{\textbf{Primary function}} \\
        \midrule

        Supervisor Agent &
        Mission state and execution events &
        High-level coordination decision &
        \RaggedRight Selects the appropriate mission response using explicit rules and event conditions \\

        Navigation and Exploration Agent &
        Pose, map, current goal, and mission state &
        Navigation actions and exploration subgoals &
        Explores the environment, approaches candidates, and reports navigation progress \\

        Object Detection Agent &
        Visual observations and target representation &
        Candidate detections &
        Identifies possible target instances without declaring mission success \\

        Target Validation Agent &
        Candidate observation and complete target description &
        Validation result &
        Determines whether an approached candidate satisfies the intended goal \\

        Semantic Advisor Agent &
        Target, recent observations, and mission context &
        High-level semantic guidance &
        \RaggedRight Suggests promising regions or search directions when conventional search is insufficient \\

        Mission Memory &
        Mission events and intermediate outcomes &
        Stored episode-level records &
        \RaggedRight Preserves information relevant to later candidate, exploration, recovery, and supervisory decisions \\

        \bottomrule
    \end{tabularx}
\end{table}

% ----------------------------------------------------------------
\subsection{Architectural Components}
% ----------------------------------------------------------------

% ----------------------------------------------------------------
\subsubsection{Supervisor Agent}
% ----------------------------------------------------------------

The Supervisor Agent is the central coordination element of F2A-ObjectNav. It monitors the mission state, receives event information from the specialized components, and selects high-level responses through explicit rules and event conditions. At mission initialization, the Supervisor establishes the default exploration, detection, and validation settings. During execution, events such as candidate detection, validation rejection, repeated movement, navigation stagnation, unsuccessful candidate approach, or insufficient search progress can cause it to revise the current mission behavior.

Depending on the detected event, the Supervisor may continue the current search, update Mission Memory, select another exploration strategy, activate recovery, clear or replace a candidate, or request semantic guidance. Importantly, the Supervisor is rule-based. It does not use a learned policy or a large language model as the primary mission controller. Language and multimodal models are instead assigned specialized roles within the architecture.

\begin{figure}[H]
    \centering
    \IfFileExists{figures/fig6.pdf}{
        \includegraphics[width=0.58\linewidth,keepaspectratio]{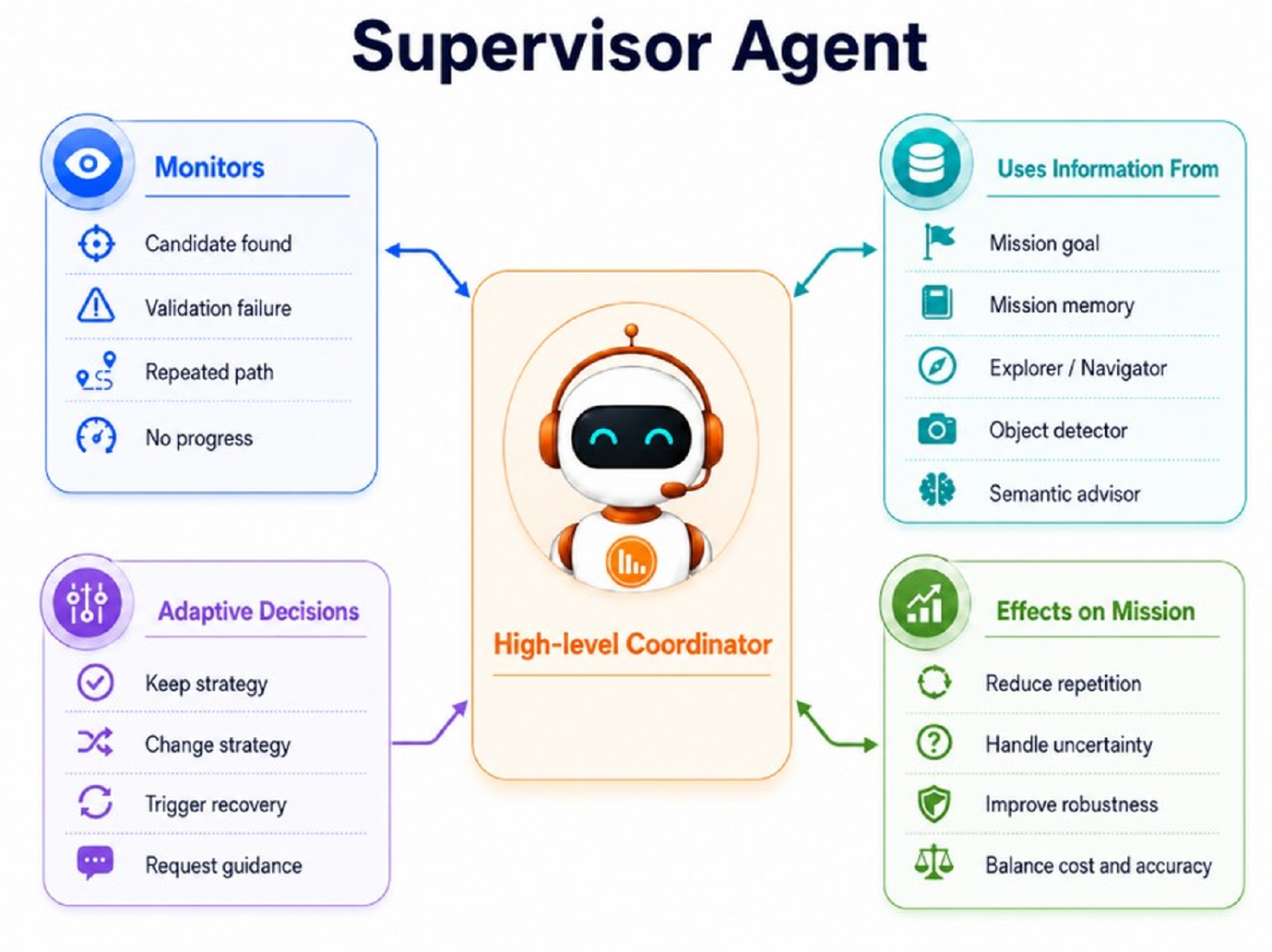}
    }{
        \fbox{\parbox[c][4.5cm][c]{0.54\linewidth}{
            \centering
            \textbf{Figure placeholder}\\[4pt]
            Supervisor Agent\\[4pt]
            \texttt{figures/fig6.pdf}
        }}
    }
    \caption{Supervisor Agent and its principal monitoring, information, and adaptive coordination roles. (AI-generated schematic using ChatGPT.)}
    \label{fig:supervisor_agent}
\end{figure}

The principal monitored events, information sources, and high-level responses of the Supervisor are summarized in Fig.~\ref{fig:supervisor_agent}.

% ----------------------------------------------------------------
\subsubsection{Navigation and Exploration Agent}
% ----------------------------------------------------------------

The Navigation and Exploration Agent is responsible for searching the environment, selecting reachable exploration goals, navigating toward them, and approaching selected target candidates. During routine exploration, it uses available pose, map, depth, and environmental information to distinguish navigable regions, obstacles, previously explored areas, and candidate exploration goals.

When the Object Detection Agent reports a candidate selected for inspection, the Navigation and Exploration Agent transitions from general exploration to candidate approach and seeks a suitable observation point near the estimated candidate location. It also monitors navigation progress and reports mission-relevant events such as repeated trajectories, failure to reach a goal, or prolonged movement without useful progress. These events can subsequently trigger replanning, recovery, strategy revision, or semantic assistance through the Supervisor.

\begin{figure}[H]
    \centering
    \IfFileExists{figures/fig7.pdf}{
        \includegraphics[width=0.4\linewidth,keepaspectratio]{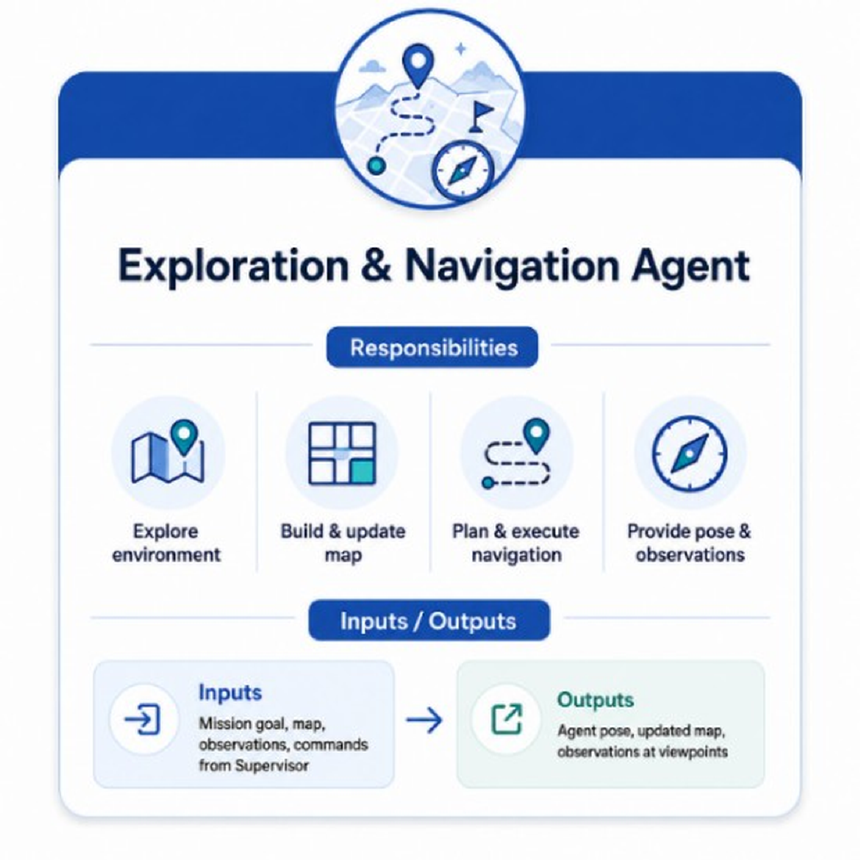}
    }{
        \fbox{\parbox[c][4.5cm][c]{0.48\linewidth}{
            \centering
            \textbf{Figure placeholder}\\[4pt]
            Navigation and Exploration Agent\\[4pt]
            \texttt{figures/fig7.pdf}
        }}
    }
    \caption{Navigation and Exploration Agent, including its main responsibilities and information flow. (AI-generated schematic using ChatGPT.)}
    \label{fig:navigation_exploration_agent}
\end{figure}

The responsibilities and principal information flow of the Navigation and Exploration Agent are summarized in Fig.~\ref{fig:navigation_exploration_agent}.

% ----------------------------------------------------------------
\subsubsection{Object Detection Agent}
% ----------------------------------------------------------------

The Object Detection Agent analyzes visual observations collected during navigation and identifies possible instances of the specified target. In the implemented architecture, YOLO11x serves as the primary detector for common object categories, while YOLO-World v8x provides open-vocabulary detection when the target representation requires greater flexibility. The detailed model configuration is described in Section~\ref{sec:experimental_methodology}.

The detector output may contain the predicted object class, confidence information, image-space localization, and an estimated candidate location in the environment. Detection alone does not constitute mission success. Each detection is treated only as a candidate and is subsequently checked against Mission Memory and, when selected, evaluated by the Target Validation Agent.

This separation between candidate detection and final target acceptance is a deliberate architectural choice. It allows perceptual proposals to be reconsidered using additional visual evidence and the complete user-specified goal rather than relying only on detector confidence.

\begin{figure}[H]
    \centering
    \IfFileExists{figures/fig8.pdf}{
        \includegraphics[width=0.4\linewidth,keepaspectratio]{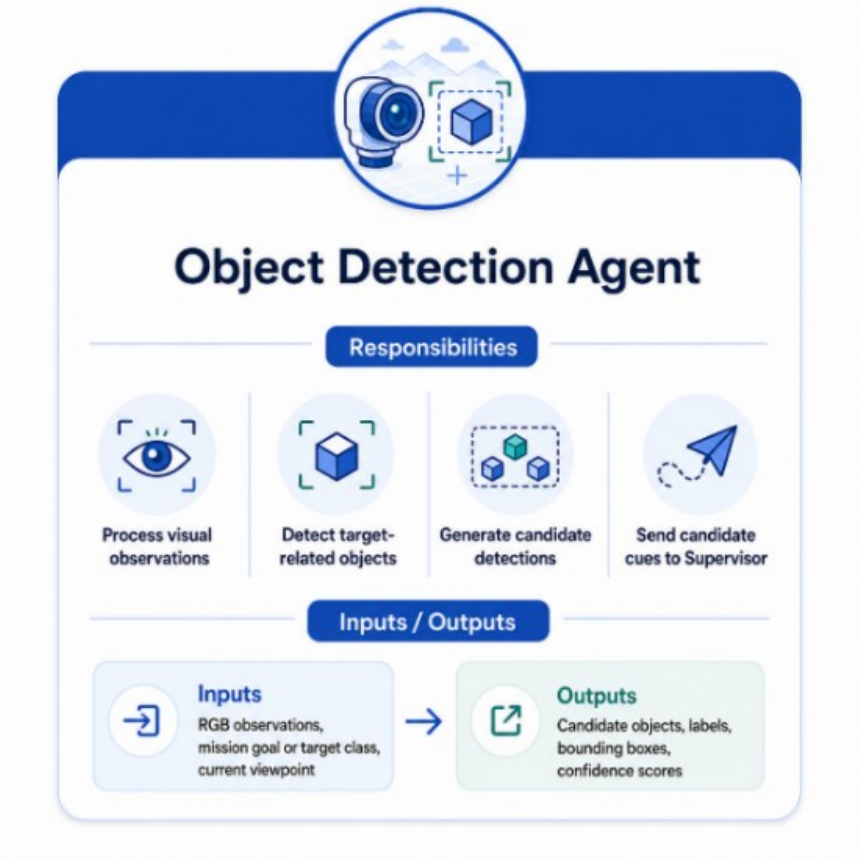}
    }{
        \fbox{\parbox[c][4.5cm][c]{0.48\linewidth}{
            \centering
            \textbf{Figure placeholder}\\[4pt]
            Object Detection Agent\\[4pt]
            \texttt{figures/fig8.pdf}
        }}
    }
    \caption{Object Detection Agent and its role in generating candidate target observations for subsequent inspection and validation. (AI-generated schematic using ChatGPT.)}
    \label{fig:object_detection_agent}
\end{figure}

The functional role and principal inputs and outputs of the Object Detection Agent are summarized in Fig.~\ref{fig:object_detection_agent}.

% ----------------------------------------------------------------
\subsubsection{Target Validation Agent}
% ----------------------------------------------------------------

The Target Validation Agent performs the final semantic assessment of an approached candidate. Unlike the Object Detection Agent, it is not executed continuously on every observation. It is activated after the Navigation and Exploration Agent reaches a viewpoint from which sufficient visual evidence about the candidate can be collected.

Validation uses the candidate observation, contextual scene information, and the complete target description. This allows the system to evaluate not only the general object category but also additional properties specified in the goal, such as color, material, subtype, or other visible constraints. In the implemented system, GPT-4.1 mini is used for this final semantic validation.

A positively validated candidate triggers mission termination, whereas a rejected candidate is recorded in Mission Memory and the system resumes exploration. When the surrounding supervisory logic determines that the available visual evidence is insufficient for reliable assessment, it can request an additional observation or a modified viewpoint before another validation attempt. The validator interface itself remains binary; additional evidence collection is handled by the surrounding architecture rather than represented as a third validation output class.

\begin{figure}[H]
    \centering
    \IfFileExists{figures/fig9.pdf}{
        \includegraphics[width=0.4\linewidth,keepaspectratio]{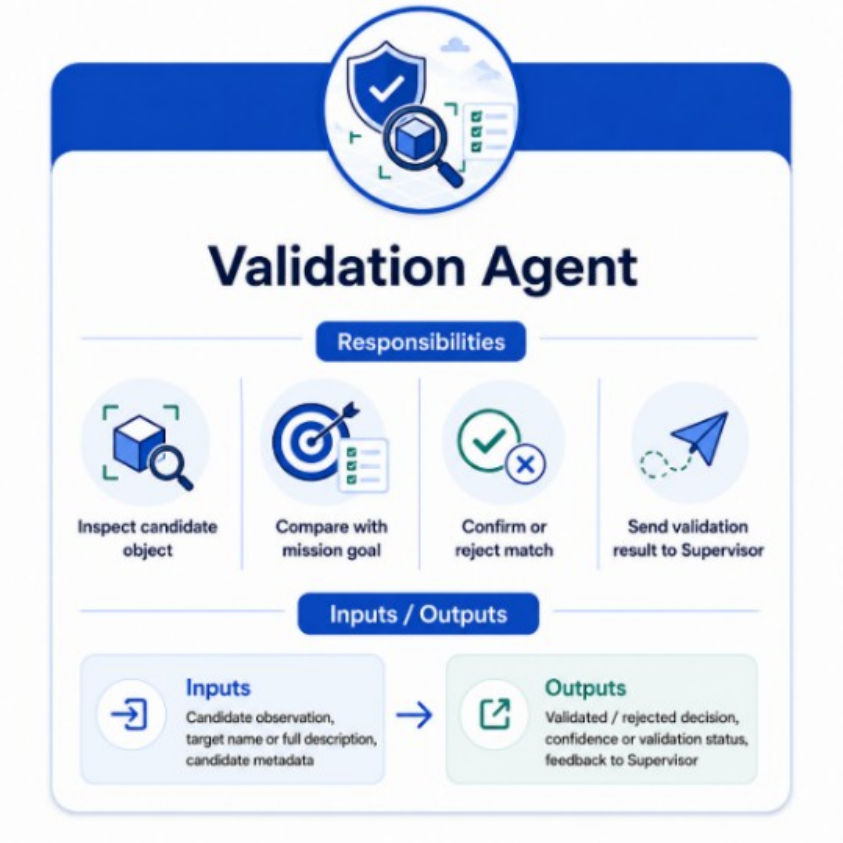}
    }{
        \fbox{\parbox[c][4.5cm][c]{0.48\linewidth}{
            \centering
            \textbf{Figure placeholder}\\[4pt]
            Target Validation Agent\\[4pt]
            \texttt{figures/fig9.pdf}
        }}
    }
    \caption{Target Validation Agent and its role in assessing approached candidates against the complete mission goal. (AI-generated schematic using ChatGPT.)}
    \label{fig:target_validation_agent}
\end{figure}

The functional role of the Target Validation Agent is summarized in Fig.~\ref{fig:target_validation_agent}.

% ----------------------------------------------------------------
\subsubsection{Semantic Advisor Agent}
% ----------------------------------------------------------------

The Semantic Advisor Agent provides high-level contextual support when the default exploration and detection process is no longer producing sufficient progress. Typical activation conditions include repeated exploration, prolonged search without a useful candidate, unsuccessful recovery, or other mission states in which semantic context may help prioritize subsequent search.

When activated, the Advisor receives the mission target together with selected recent observations and relevant mission context. GPT-4o is used to infer a promising region, direction, or contextual relation associated with the target. The returned suggestion is not used as a direct low-level control command. Instead, it is interpreted by the Supervisor and converted by the Navigation and Exploration Agent into a feasible exploration objective.

Semantic reasoning is therefore treated as a conditional resource rather than a continuously active navigation controller.

\begin{figure}[H]
    \centering
    \IfFileExists{figures/fig10.pdf}{
        \includegraphics[width=0.4\linewidth,keepaspectratio]{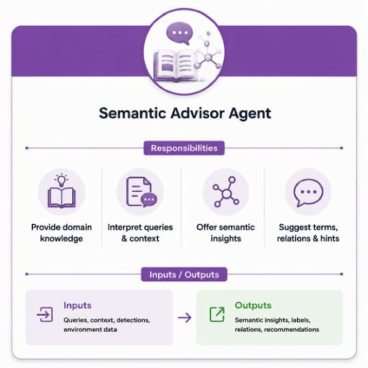}
    }{
        \fbox{\parbox[c][4.5cm][c]{0.48\linewidth}{
            \centering
            \textbf{Figure placeholder}\\[4pt]
            Semantic Advisor Agent\\[4pt]
            \texttt{figures/fig10.pdf}
        }}
    }
    \caption{Semantic Advisor Agent, which provides conditional high-level contextual guidance without directly controlling low-level navigation. (AI-generated schematic using ChatGPT.)}
    \label{fig:semantic_advisor_agent}
\end{figure}

The Semantic Advisor and its high-level advisory role are summarized in Fig.~\ref{fig:semantic_advisor_agent}.

% ----------------------------------------------------------------
\subsubsection{Mission Memory}
% ----------------------------------------------------------------

Mission Memory stores episode-level information that may influence subsequent decisions during the same mission. Its records can include visited or repeatedly explored regions, rejected candidates, recent navigation goals, unsuccessful candidate-approach attempts, recovery events, strategy changes, and previous semantic-guidance requests.

When a candidate is rejected, its estimated location and associated information are stored so that subsequent detections corresponding to the same instance can be filtered. Information about repeated trajectories or ineffective regions can likewise be used by the Supervisor when deciding whether to continue the current strategy, initiate recovery, or select another exploration behavior.

Mission Memory is updated in response to meaningful mission events rather than after every low-level action. Its purpose is therefore not to maintain a complete history of all observations, but to preserve information that can reduce redundant behavior and support later supervisory decisions.

\begin{figure}[H]
    \centering
    \IfFileExists{figures/fig11.pdf}{
        \includegraphics[width=0.4\linewidth,keepaspectratio]{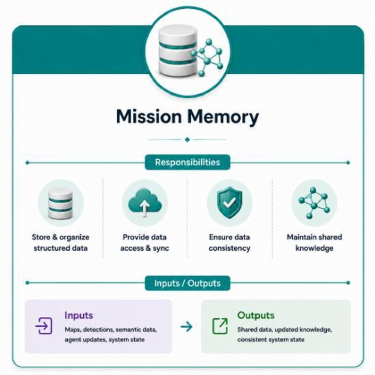}
    }{
        \fbox{\parbox[c][4.5cm][c]{0.48\linewidth}{
            \centering
            \textbf{Figure placeholder}\\[4pt]
            Mission Memory\\[4pt]
            \texttt{figures/fig11.pdf}
        }}
    }
    \caption{Mission Memory and its role in preserving mission-relevant information for later supervisory decisions. (AI-generated schematic using ChatGPT.)}
    \label{fig:mission_memory}
\end{figure}

The information-management role of Mission Memory is summarized in Fig.~\ref{fig:mission_memory}.

% ----------------------------------------------------------------
\subsection{Adaptive Mechanisms and Trigger Logic}
% ----------------------------------------------------------------

The six architectural elements are connected through a set of adaptive mechanisms that determine how the system responds to mission-level events. These mechanisms do not operate continuously. Instead, they are activated when the Supervisor detects conditions requiring a departure from the default execution path.

Table~\ref{tab:adaptive_mechanisms} summarizes the principal adaptive mechanisms and their typical activation conditions.

\begin{table}[!ht]
    \centering
    \caption{Adaptive mechanisms used in F2A-ObjectNav.}
    \label{tab:adaptive_mechanisms}
    \footnotesize
    \ArchitectureTableStyle
    \begin{tabularx}{\linewidth}{C{0.22\linewidth}C{0.30\linewidth}Z}
        \toprule
        \rowcolor{ArchitectureHeader}
        \textcolor{TableHeaderText}{\textbf{Adaptive mechanism}} &
        \textcolor{TableHeaderText}{\textbf{Typical activation condition}} &
        \textcolor{TableHeaderText}{\textbf{Role in the architecture}} \\
        \midrule

        Strategy switching &
        Insufficient progress, repeated failure, or ineffective current exploration behavior &
        \RaggedRight Revises the active exploration strategy using the alternatives available to the Supervisor \\

        Semantic-guided exploration &
        \RaggedRight Prolonged search, repeated exploration, unsuccessful recovery, or insufficient semantic evidence &
        \RaggedRight Requests high-level contextual guidance about a promising region or direction without directly controlling robot motion \\

        Candidate management &
        Candidate detection, rejection, loss, or re-observation &
        \RaggedRight Maintains candidate status and uses Mission Memory to avoid repeated evaluation of rejected instances \\

        Spatial candidate merging &
        Multiple detections correspond to approximately the same estimated location &
        Consolidates repeated observations into a single candidate record \\

        Recovery &
        \RaggedRight Navigation stagnation, repeated trajectory, or failure to reach the current navigation objective &
        \RaggedRight Replans, changes viewpoint or subgoal, clears an ineffective candidate, or resumes exploration \\

        Goal-aware validation &
        The agent reaches a candidate with sufficient visual evidence &
        \RaggedRight Evaluates the candidate against the complete target description before mission termination \\

        \bottomrule
    \end{tabularx}
\end{table}

The recorded executions include switches between frontier-based and random exploration after ineffective search, together with recovery actions such as replanning, scanning, and relocalization. Rejected candidates are retained in Mission Memory for subsequent decisions within the same episode; memory is reset before the next episode. Table~\ref{tab:adaptive_mechanisms} summarizes the event--response relationships.

The trigger logic is intentionally separated from low-level perception and motion control. This design allows straightforward missions to proceed through the default exploration--detection--navigation--validation sequence, while additional mechanisms are introduced only when the mission state indicates that they are required.

% ----------------------------------------------------------------
\subsection{Adaptive Decision-Making Flow}
% ----------------------------------------------------------------

F2A-ObjectNav executes each mission as a repeated perception--decision--action cycle in which the active behavior can be revised according to the current mission state. At initialization, the user goal is converted into a primary object representation for detection and a complete target description for validation. Mission Memory is initialized, and the Supervisor activates the default exploration and detection process.

During normal execution, the Navigation and Exploration Agent moves through the environment while the Object Detection Agent processes incoming observations. If no target candidate is detected and useful exploration progress continues, the active strategy remains unchanged. When the Supervisor detects repeated exploration, navigation stagnation, failure to reach the current goal, or insufficient progress, it evaluates the recent mission state and selects an appropriate adaptive response.

When a candidate is detected, its estimated location and identity are first compared with previously rejected candidates stored in Mission Memory. Candidates corresponding to rejected instances are filtered to avoid unnecessary re-evaluation. Otherwise, the Navigation and Exploration Agent approaches the candidate and obtains a suitable observation for validation. The Target Validation Agent then evaluates the candidate against the complete target description.

A positively validated candidate triggers mission termination. If the candidate is rejected, its record is added to Mission Memory and exploration resumes. If the available observation does not provide sufficient evidence for a reliable validation decision, an additional viewpoint or visual observation can be requested before validation is repeated. Multiple detections corresponding to approximately the same physical location are consolidated through spatial candidate merging.

If the agent cannot reach a selected candidate or navigation progress deteriorates, the recovery mechanism is activated. Depending on the detected condition, recovery may include path replanning, selecting another viewpoint or exploration goal, clearing the current candidate, or returning to general exploration. If repeated recovery or strategy revision does not restore useful progress, the Supervisor may invoke the Semantic Advisor.

The Semantic Advisor provides only high-level contextual guidance, such as a likely region or search direction. The Supervisor determines how this information should influence the mission, and the Navigation and Exploration Agent converts the resulting decision into an executable navigation objective. Consequently, semantic guidance supplements rather than replaces the conventional navigation process.

The overall supervisory sequence can be summarized as follows:

\begin{enumerate}
    \item Interpret the mission goal and initialize Mission Memory.
    \item Execute the default exploration and detection process.
    \item Retain the current behavior while useful mission progress continues.
    \item When a candidate is detected, compare it with Mission Memory and approach it if it has not been rejected previously.
    \item Validate the candidate using the complete target description.
    \item On rejection, update Mission Memory and resume exploration.
    \item On navigation stagnation or failure, activate recovery or revise the exploration strategy.
    \item If conventional recovery remains ineffective, request conditional semantic guidance.
    \item Continue until the system accepts a candidate as the final target or the execution-step budget is exhausted.
\end{enumerate}

These mechanisms allow F2A-ObjectNav to retain a simple default execution path for straightforward missions while introducing memory, recovery, strategy revision, and semantic reasoning only when the evolving mission state requires additional intervention.

% ================================================================
\section{Experimental Methodology}
\label{sec:experimental_methodology}
% ================================================================

% ----------------------------------------------------------------
\subsection{Evaluation Environments}
\label{subsec:evaluation_environments}
% ----------------------------------------------------------------

The proposed adaptive architecture is evaluated in three indoor simulation environments: AI2-THOR \cite{kolve2017ai2thor}, RoboTHOR \cite{deitke2020robothor}, and ProcTHOR \cite{deitke2022procthor}. These environments provide complementary settings for examining the same architecture under different scene structures, navigation interfaces, and object arrangements. AI2-THOR contains interactive residential scenes, RoboTHOR emphasizes embodied navigation under robot-oriented motion and perception constraints, and ProcTHOR provides procedurally generated houses with greater variation in layout, room connectivity, and object placement.

The purpose of using multiple simulation environments is not to rank them by difficulty, but to examine whether the same high-level supervisory logic and adaptive mechanisms can operate across different scene structures and search conditions. The high-level architecture, component roles, supervisory logic, and adaptive mechanisms are retained across the three environments. Only simulator-specific interfaces for observations, pose information, and executable actions are adapted to the corresponding platform.

\begin{figure}[H]
    \centering
    \IfFileExists{figures/fig12.pdf}{
        \includegraphics[width=0.72\linewidth,keepaspectratio]{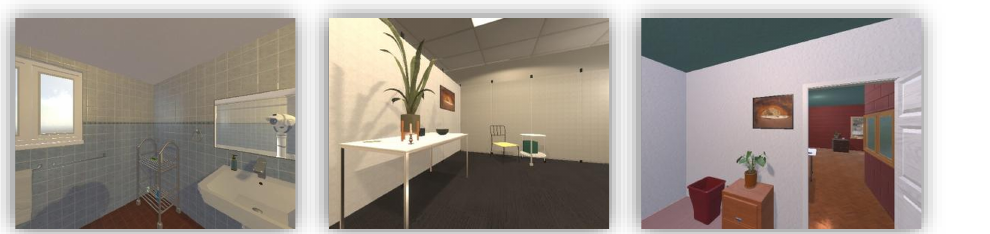}
    }{
        \fbox{\parbox[c][4.6cm][c]{0.92\linewidth}{
            \centering
            \textbf{Figure placeholder}\\[4pt]
            Evaluation environments:\\
            AI2-THOR, RoboTHOR, and ProcTHOR\\[4pt]
            \texttt{figures/fig12.pdf}
        }}
    }
    \caption{Representative scenes from the three simulation environments used for evaluation: AI2-THOR, RoboTHOR, and ProcTHOR.}
    \label{fig:evaluation_environments}
\end{figure}

Representative scenes from the three evaluation environments are shown in Fig.~\ref{fig:evaluation_environments}. Additional environment characteristics and implementation settings are summarized in Appendix~\ref{app:experimental_settings}.

\begin{figure}[H]
    \centering
    \IfFileExists{figures/fig13.pdf}{
        \includegraphics[width=0.62\linewidth,keepaspectratio]{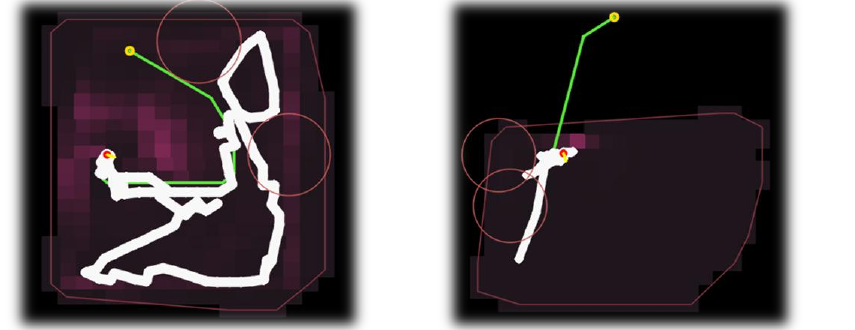}
    }{
        \fbox{\parbox[c][4.2cm][c]{0.78\linewidth}{
            \centering
            \textbf{Figure placeholder}\\[4pt]
            Representative navigation maps generated during evaluation\\[4pt]
            \texttt{figures/fig13.pdf}
        }}
    }
    \caption{Examples of generated maps during simulation, including the agent trajectory, intermediate goals, explored regions, and related navigation information.}
    \label{fig:evaluation_maps}
\end{figure}

Representative navigation maps generated during evaluation are shown in Fig.~\ref{fig:evaluation_maps}. They illustrate recorded trajectories, intermediate navigation goals, explored regions, and related navigation information produced during mission execution.

% ----------------------------------------------------------------
\subsection{Models and System Configuration}
\label{subsec:model_configuration}
% ----------------------------------------------------------------

At the beginning of each episode, the mission goal is provided as an object category or a natural-language description. GPT-4o mini is used for goal interpretation \cite{openai2024gpt4omini}. It separates the primary object representation used for detection from the complete description retained for final validation.

\begin{figure}[H]
    \centering
    \IfFileExists{figures/fig14.pdf}{
        \includegraphics[width=0.72\linewidth,keepaspectratio]{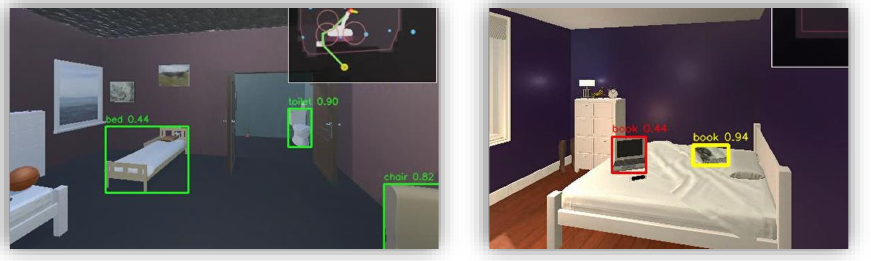}
    }{
        \fbox{\parbox[c][4.2cm][c]{0.82\linewidth}{
            \centering
            \textbf{Figure placeholder}\\[4pt]
            Representative object-detection outputs\\[4pt]
            \texttt{figures/fig14.pdf}
        }}
    }
    \caption{Example of the object-detection process during simulation, showing representative YOLO11x (left) and YOLO-World v8x (right) outputs.}
    \label{fig:detection_examples}
\end{figure}

YOLO11x is used as the primary detector for common object categories \cite{khanam2024yolov11}, while YOLO-World v8x provides open-vocabulary detection when the target representation requires more flexible class names or aliases \cite{cheng2024yoloworld}. Representative outputs from the two detection modes are shown in Fig.~\ref{fig:detection_examples}.

\begin{figure}[H]
    \centering
    \IfFileExists{figures/fig15.pdf}{
        \includegraphics[width=0.72\linewidth,keepaspectratio]{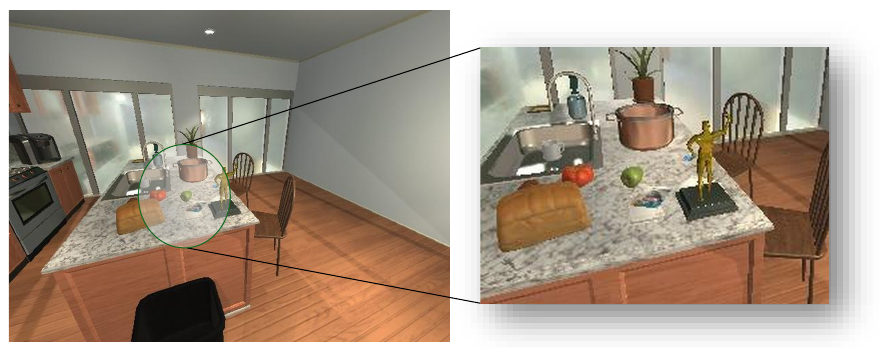}
    }{
        \fbox{\parbox[c][4.4cm][c]{0.82\linewidth}{
            \centering
            \textbf{Figure placeholder}\\[4pt]
            Representative target-validation example\\[4pt]
            \texttt{figures/fig15.pdf}
        }}
    }
    \caption{Example of the target-object validation process; the illustrated target object is an apple.}
    \label{fig:validation_example}
\end{figure}

GPT-4.1 mini is used for final candidate validation \cite{openai2025gpt41}. A representative validation example is shown in Fig.~\ref{fig:validation_example}; the validator evaluates a focused candidate observation together with wider scene context and the complete target specification.

GPT-4o is assigned to the Semantic Advisor and is invoked only when the rule-based Supervisor determines that additional semantic guidance is required \cite{openai2024gpt4o}.

The functional role of each model is kept fixed across AI2-THOR, RoboTHOR, and ProcTHOR. Table~\ref{tab:model_roles} summarizes the model assignments and their activation modes.

\begin{table}[!ht]
    \centering
    \caption{Models, roles, and activation modes used in F2A-ObjectNav.}
    \label{tab:model_roles}
    \footnotesize
    \ArchitectureTableStyle
    \begin{tabularx}{\linewidth}{C{0.24\linewidth}C{0.19\linewidth}C{0.37\linewidth}Z}
        \toprule
        \rowcolor{ArchitectureHeader}
        \textcolor{TableHeaderText}{\textbf{Component}} &
        \textcolor{TableHeaderText}{\textbf{Model}} &
        \textcolor{TableHeaderText}{\textbf{Role}} &
        \textcolor{TableHeaderText}{\textbf{Activation}} \\
        \midrule

        Goal interpretation &
        GPT-4o mini &
        \RaggedRight Extracts the primary target representation while preserving the complete description for validation &
        Mission initialization \\

        Primary object detection &
        YOLO11x &
        Detects common object categories during routine perception &
        During normal search when applicable \\

        Open-vocabulary detection &
        YOLO-World v8x &
        \RaggedRight Detects target classes using flexible object names and accepted aliases &
        When open-vocabulary detection is required \\

        Target validation &
        GPT-4.1 mini &
        Evaluates an approached candidate against the complete target description &
        After candidate approach \\

        Semantic guidance &
        GPT-4o &
        \RaggedRight Provides high-level contextual guidance about promising search regions or directions &
        On request from the Supervisor \\
        \bottomrule
    \end{tabularx}
\end{table}

The GPT-based components use structured prompts and outputs to maintain predictable interactions with the supervisory logic. The prompts used for goal interpretation, semantic guidance, and target validation are documented in Appendix~\ref{app:prompts}. YOLO11x does not receive a free-form natural-language prompt and operates using its supported object classes together with the alias mapping maintained by the system. YOLO-World v8x receives a controlled list of target classes and equivalent object names rather than an unrestricted natural-language instruction.

% ----------------------------------------------------------------
\subsection{Episode Protocol and Termination}
\label{subsec:episode_protocol}

Stage II comprised 25 episodes in each of AI2-THOR, RoboTHOR, and ProcTHOR, for 75 episodes in total. Scenes were selected randomly from the available scene collection of each environment. These are study-specific evaluations in the three simulation environments, rather than evaluations over the complete official benchmark episode sets.

At the beginning of each episode, the environment and mission state were initialized, the agent was placed at a valid starting pose, and Mission Memory was reset. The target was specified by an object category or a natural-language description and was present in the scene, but its location was not supplied to the agent. The Target Interpreter extracted a detector-facing category and retained the full description for final validation. Mission execution then proceeded without manual intervention.

The rule-based Supervisor coordinated navigation, detection, candidate management, recovery, and conditional semantic guidance. When a candidate was selected, the agent approached an observation position and the Target Validation Agent assessed the candidate against the complete user request. Rejection updated Mission Memory and returned the system to search; it did not itself terminate the episode.

\paragraph{Termination and outcome assessment.}
The system terminated when it accepted a candidate as the final target or exhausted its execution-step budget. System acceptance and evaluation success were distinguished: successful episodes required a stopping decision, a visible target within 1~m Euclidean distance of the agent's camera, and a correct match to the requested target. Target correctness was assessed by human inspection. Acceptance of an incorrect object was scored as failure, as was budget exhaustion without successful completion. The evaluation yielded 21, 10, and 11 successful episodes in AI2-THOR, RoboTHOR, and ProcTHOR, respectively, corresponding to SR values of 84.00\%, 40.00\%, and 44.00\%.

\paragraph{Execution-step convention.}
A recorded execution step denotes an operation performed during mission execution, such as forward motion, rotation, object detection, or a recovery operation. It is therefore not restricted to a translational or rotational simulator action. The same step convention is used for the episode budget and the step-normalized behavioral metrics. The budget-limited examples in Appendix~\ref{app:execution_evidence} contain approximately 1,000 recorded steps. Spatial path length for SPL is distinct from this operation count.

% ----------------------------------------------------------------
\subsection{Evaluation Metrics}
\label{subsec:evaluation_metrics}
% ----------------------------------------------------------------

The evaluation considers both the final outcome of the ObjectNav mission and the internal behavior of the proposed adaptive architecture. To distinguish their roles clearly, the evaluation measures are organized into three groups: standard ObjectNav metrics, system-level performance measures, and architecture-specific behavioral metrics.

The standard ObjectNav metrics are Success Rate (SR) and Success-weighted Path Length (SPL). SR measures the proportion of successfully completed missions, while SPL evaluates mission success together with path efficiency according to the standard embodied-navigation formulation \cite{anderson2018evaluation}. These two measures provide the primary basis for assessing navigation effectiveness.

The system-level measures are Task Time (TT), Validation Accuracy (VA), and API Cost (AC). TT measures the average duration of an episode from mission initialization to termination. VA evaluates the correctness of candidate-validation decisions. API Cost measures the average model-related inference cost associated with externally accessed models during mission execution. These measures complement SR and SPL by characterizing execution time, candidate-level validation correctness, and external inference expenditure. API Cost does not measure total computational cost.

In addition to these task- and system-level measures, five architecture-specific behavioral metrics are used to characterize how the supervisory coordination mechanism operates during execution. Semantic Guidance Rate (SGR) measures the frequency with which the Semantic Advisor is invoked relative to episode length. Strategy Switching Rate (SSR) measures the frequency of revisions to the active navigation or exploration strategy. Recovery Activation combines the occurrence of navigation problems with the normalized recovery response within affected episodes. Memory-Assisted Loop Response measures recorded memory-supported corrective actions associated with detected loops or repeated states rather than total memory accesses or a causal contribution to escape success. Decision Stability characterizes the frequency of high-level decision revisions relative to episode length and should be interpreted together with mission outcome rather than as an independent measure of navigation quality.

Table~\ref{tab:evaluation_metrics} summarizes the complete evaluation set. The exact mathematical definitions, normalization procedures, and evaluation-level aggregation rules are provided in Appendix~\ref{app:metric_definitions}. All metrics are calculated separately for AI2-THOR, RoboTHOR, and ProcTHOR.

\begin{table}[!ht]
    \centering
    \caption{Metrics used to evaluate F2A-ObjectNav.}
    \label{tab:evaluation_metrics}
    \footnotesize
    \EvaluationTableStyle
    \begin{tabularx}{\linewidth}{C{0.22\linewidth}C{0.28\linewidth}Z}
        \toprule
        \rowcolor{EvaluationHeader}
        \textcolor{TableHeaderText}{\textbf{Group}} & \textcolor{TableHeaderText}{\textbf{Metric}} & \textcolor{TableHeaderText}{\textbf{Purpose}} \\
        \midrule

        Standard ObjectNav &
        Success Rate (SR) &
        Measures successful mission completion. \\

        Standard ObjectNav &
        Success-weighted Path Length (SPL) &
        Measures mission success together with path efficiency. \\

        \midrule

        System-level &
        Task Time (TT) &
        Measures average episode duration from initialization to termination. \\

        System-level &
        Validation Accuracy (VA) &
        Measures the correctness of candidate-validation decisions. \\

        System-level &
        API Cost (AC) &
        \RaggedRight Measures the average inference cost of externally accessed models per episode. \\

        \midrule

        Architecture-specific &
        Semantic Guidance Rate (SGR) &
        \RaggedRight Measures the frequency of Semantic Advisor activation relative to episode length. \\

        Architecture-specific &
        Strategy Switching Rate (SSR) &
        Measures the frequency of navigation or exploration strategy revisions. \\

        Architecture-specific &
        Recovery Activation &
        Measures recovery responses to detected navigation problems. \\

        Architecture-specific &
        Memory-Assisted Loop Response &
        \RaggedRight Measures memory-supported responses to detected navigation loops or repeated states. \\

        Architecture-specific &
        Decision Stability &
        \RaggedRight Measures the relative stability of high-level supervisory decisions during execution. \\

        \bottomrule
    \end{tabularx}
\end{table}

% ================================================================
\section{Results}
% ================================================================

% ----------------------------------------------------------------
\subsection{Task- and System-Level Performance}
% ----------------------------------------------------------------

Table~\ref{tab:main_results} summarizes the task- and system-level performance of F2A-ObjectNav in AI2-THOR, RoboTHOR, and ProcTHOR. AI2-THOR produced the highest Success Rate (SR) and Success-weighted Path Length (SPL), together with the shortest average Task Time (TT) and the lowest API Cost among the three evaluated environments. RoboTHOR and ProcTHOR produced lower SR values and longer execution times under the corresponding evaluation protocols.

\begin{table}[!ht]
    \centering
    \caption{Task- and system-level performance of F2A-ObjectNav across the three evaluation environments.}
    \label{tab:main_results}
    \footnotesize
    \EvaluationTableStyle
    \begin{tabularx}{\linewidth}{C{0.15\linewidth}*{5}{Z}}
        \toprule
        \rowcolor{EvaluationHeader}
        \textcolor{TableHeaderText}{\textbf{Environment}} &
        \textcolor{TableHeaderText}{\textbf{SR (\%)}} &
        \textcolor{TableHeaderText}{\textbf{SPL}} &
        \textcolor{TableHeaderText}{\textbf{TT (s)}} &
        \textcolor{TableHeaderText}{\textbf{VA (\%)}} &
        \textcolor{TableHeaderText}{\textbf{API Cost (\$)}} \\
        \midrule
        AI2-THOR & 84.00 & 0.52 & 48.63 & 89.5 & 0.0125 \\
        RoboTHOR & 40.00 & 0.22 & 101.04 & 86.7 & 0.0273 \\
        ProcTHOR & 44.00 & 0.37 & 109.73 & 84.2 & 0.0294 \\
        \bottomrule
    \end{tabularx}
\end{table}

Validation Accuracy (VA) varied within a narrower range, from 84.2\% to 89.5\%, than the corresponding SR values. VA and SR have different denominators: VA concerns evaluated candidate decisions, whereas SR concerns complete episodes. Their numerical difference alone cannot identify the dominant failure stage. The unsuccessful candidate-acquisition trace in Appendix~\ref{app:execution_evidence} provides a concrete example of failure before final validation.

RoboTHOR and ProcTHOR also produced higher average Task Time and API Cost than AI2-THOR. These differences are consistent with longer mission execution and greater use of externally accessed model-dependent operations in the recorded runs. However, the aggregate results do not isolate the independent contribution of any individual component or adaptive mechanism to these differences.

\begin{figure}[H]
    \centering
    \IfFileExists{figures/fig16.pdf}{
        \includegraphics[width=0.98\linewidth,keepaspectratio]{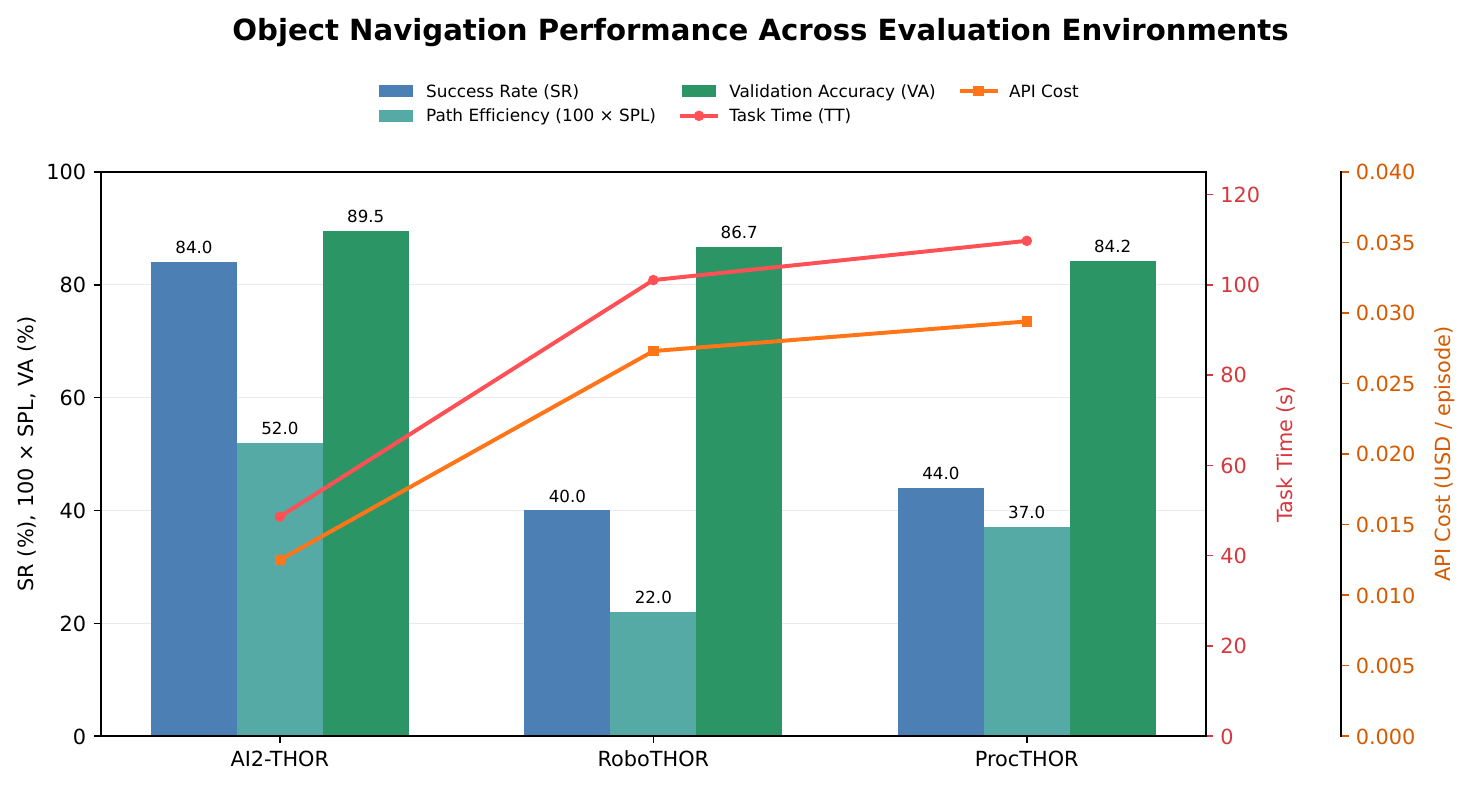}
    }{
        \fbox{\parbox[c][4.6cm][c]{0.84\linewidth}{
            \centering
            \textbf{Figure placeholder}\\[4pt]
            Graphical comparison of task- and system-level metrics\\[4pt]
            \texttt{figures/fig16.pdf}
        }}
    }
    \caption{Task- and system-level results shown as grouped bars and lines. Bars show SR, $100\times$SPL, and VA on the left axis. The red line shows Task Time on the inner right axis, and the orange line shows API Cost on the outer right axis. Values correspond to Table~\ref{tab:main_results}; API Cost is expressed in USD per episode.}
    \label{fig:standard_metrics_results}
\end{figure}

Figure~\ref{fig:standard_metrics_results} provides the corresponding graphical summary of the reported task- and system-level results.

% ----------------------------------------------------------------
\subsection{Architecture-Specific Behavioral Analysis}
% ----------------------------------------------------------------

SR and SPL characterize the final navigation outcome, while the system-level measures describe execution time, validation reliability, and model-related cost. They do not, however, reveal how frequently the adaptive coordination mechanisms are used during execution. Table~\ref{tab:agent_metrics} therefore reports the architecture-specific behavioral metrics defined for F2A-ObjectNav.

\begin{table}[!ht]
    \centering
    \caption{Architecture-specific behavioral metrics of F2A-ObjectNav across the three evaluation environments.}
    \label{tab:agent_metrics}
    \footnotesize
    \EvaluationTableStyle
    \begin{tabularx}{\linewidth}{C{0.15\linewidth}*{5}{Z}}
        \toprule
        \rowcolor{EvaluationHeader}
        \textcolor{TableHeaderText}{\textbf{Environment}} &
        \textcolor{TableHeaderText}{\textbf{SGR (\%)}} &
        \textcolor{TableHeaderText}{\textbf{SSR (\%)}} &
        \textcolor{TableHeaderText}{\textbf{Recovery (\%)}} &
        \textcolor{TableHeaderText}{\textbf{Memory (\%)}} &
        \textcolor{TableHeaderText}{\textbf{Stability (\%)}} \\
        \midrule
        AI2-THOR & 0.06 & 0.61 & 13.3 & 0.80 & 98.90 \\
        RoboTHOR & 0.36 & 4.01 & 26.8 & 15.00 & 92.80 \\
        ProcTHOR & 0.65 & 5.20 & 30.6 & 7.30 & 90.50 \\
        \bottomrule
    \end{tabularx}
\end{table}

AI2-THOR produced the lowest Semantic Guidance Rate (SGR), Strategy Switching Rate (SSR), and Recovery Activation, together with the highest Decision Stability. Under the metric definitions adopted in this study, this pattern indicates that fewer high-level interventions were recorded relative to episode execution in AI2-THOR.

RoboTHOR exhibited higher rates of strategy revision and recovery than AI2-THOR and produced the highest Memory-Assisted Loop Response value among the three environments. Under the definition used in this work, this value reflects a greater proportion of memory-supported responses to detected navigation loops or repeated states rather than the total number of Mission Memory accesses.

ProcTHOR produced the highest SGR, SSR, and Recovery Activation values. The low absolute SGR values across all three environments are consistent with the intended role of the Semantic Advisor as a conditional resource rather than a continuously active controller. Strategy switching also remained relatively infrequent with respect to the total number of recorded episode steps.

\begin{figure}[H]
    \centering
    \IfFileExists{figures/fig17.pdf}{
        \includegraphics[width=0.98\linewidth,keepaspectratio]{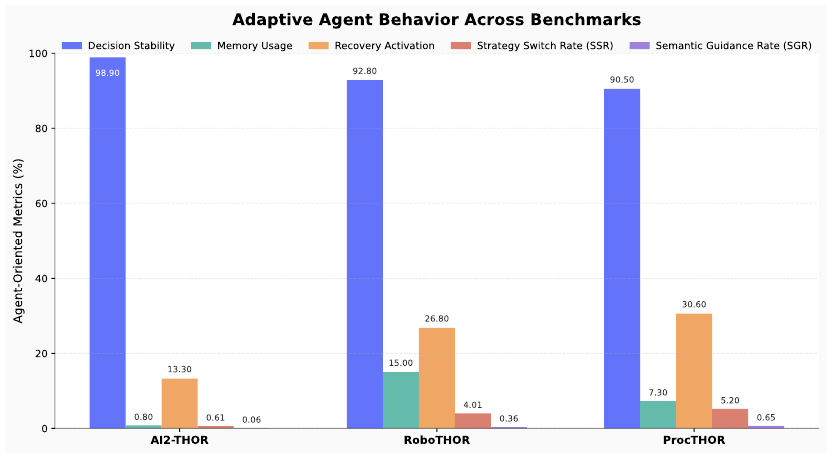}
    }{
        \fbox{\parbox[c][4.8cm][c]{0.90\linewidth}{
            \centering
            \textbf{Figure placeholder}\\[4pt]
            Comparison of architecture-specific behavioral metrics across\\
            AI2-THOR, RoboTHOR, and ProcTHOR\\[4pt]
            \texttt{figures/fig17.pdf}
        }}
    }
    \caption{Comparison of the architecture-specific behavioral metrics across the three evaluation environments. The legend label ``Memory Usage'' denotes Memory-Assisted Loop Response, as defined in Appendix~\ref{app:metric_definitions}.}
    \label{fig:agent_centric_results}
\end{figure}

Taken together, these measurements show that the adaptive mechanisms were activated with different frequencies across the evaluated environments. SGR and SSR were higher in RoboTHOR and ProcTHOR than in AI2-THOR. The recovery and memory-response indicators were also higher, but these episode-averaged ratios combine event prevalence with response behavior and should not be read as raw event counts. These results characterize the operational behavior of the architecture but do not establish the causal contribution of any individual adaptive mechanism to SR, SPL, or overall mission success. Isolating such contributions would require a controlled ablation study.

% ----------------------------------------------------------------
\subsection{Execution-Trace Analysis of Adaptive Mechanisms}
\label{subsec:ablation_analysis}
% ----------------------------------------------------------------

The principal adaptive mechanisms of F2A-ObjectNav are designed to become active only when specific mission conditions require additional intervention. Consequently, their functional role cannot be fully characterized by final SR and SPL values alone. To complement the aggregate evaluation, an execution-trace analysis is therefore performed using the behavioral measurements and recorded execution traces collected during the integrated system experiments.

Unlike a conventional removal-based ablation, the evaluation was not repeated after individually disabling each component. Instead, the present analysis examines naturally occurring differences in mechanism activation across episodes and environments, including cases in which individual mechanisms remain inactive and cases in which they are repeatedly invoked. The analysis therefore addresses three questions: whether each mechanism is functionally active in the integrated architecture, whether it is triggered under the mission conditions for which it was designed, and how its activation affects the subsequent execution flow.

Table~\ref{tab:ablation_analysis} combines evaluation-wide behavioral measurements with representative execution-trace evidence for Mission Memory, recovery, strategy switching, conditional semantic guidance, candidate management, and supervisory coordination.

\begin{table}[!ht]
    \centering
    \caption{Execution-trace analysis of the principal adaptive mechanisms using evaluation-level behavioral measurements and recorded execution traces.}
    \label{tab:ablation_analysis}
    \scriptsize
    \AblationTableStyle
    \begin{tabularx}{\linewidth}{C{0.18\linewidth}C{0.25\linewidth}C{0.31\linewidth}Z}
        \toprule
        \rowcolor{AblationHeader}
        \textcolor{TableHeaderText}{\textbf{Mechanism}} &
        \textcolor{TableHeaderText}{\textbf{Aggregate diagnostic evidence}} &
        \textcolor{TableHeaderText}{\textbf{Execution-trace evidence}} &
        \textcolor{TableHeaderText}{\textbf{Observed functional role}} \\
        \midrule

        Mission Memory &
        \RaggedRight Memory-Assisted Loop Response was 0.80\%, 15.00\%, and 7.30\% in AI2-THOR, RoboTHOR, and ProcTHOR, respectively. &
        \RaggedRight In the recorded laptop mission, an incorrect candidate was rejected, stored in Mission Memory, and excluded from subsequent consideration before a later candidate was successfully confirmed. &
        \RaggedRight Preserves episode-level information so that previous candidate and navigation outcomes can influence later decisions. \\

        Recovery &
        \RaggedRight Recovery Activation increased from 13.3\% in AI2-THOR to 26.8\% in RoboTHOR and 30.6\% in ProcTHOR. &
        \RaggedRight In the difficult remote-control mission, recovery was activated repeatedly after repeated states and navigation difficulties. &
        \RaggedRight Provides event-triggered responses to stagnation, failed progress, and ineffective navigation states. \\

        Strategy switching &
        \RaggedRight SSR increased from 0.61\% in AI2-THOR to 4.01\% in RoboTHOR and 5.20\% in ProcTHOR. &
        \RaggedRight The recorded remote-control mission included switching between frontier-based and random exploration when the current search behavior became ineffective. &
        \RaggedRight Prevents the architecture from remaining indefinitely committed to one exploration strategy when progress deteriorates. \\

        Semantic Advisor &
        \RaggedRight SGR remained low at 0.06\%, 0.36\%, and 0.65\% in AI2-THOR, RoboTHOR, and ProcTHOR, respectively. &
        \RaggedRight Semantic guidance was not required in the direct microwave-success mission but was invoked repeatedly in the more difficult remote-control execution. &
        \RaggedRight Acts as a conditional source of high-level contextual guidance rather than a continuously active controller. \\

        Candidate management and validation &
        \RaggedRight Validation Accuracy remained between 84.2\% and 89.5\% across the three evaluation environments. &
        \RaggedRight In the laptop mission, an initial candidate was rejected and stored before exploration continued and a later candidate was successfully confirmed. &
        \RaggedRight Separates candidate generation from final target acceptance and prevents detector output from being treated directly as mission completion. \\

        Supervisory coordination &
        \RaggedRight Decision Stability was 98.90\%, 92.80\%, and 90.50\% in AI2-THOR, RoboTHOR, and ProcTHOR, respectively. &
        \RaggedRight Straightforward episodes proceeded with little intervention, whereas difficult executions involved recovery, strategy revision, memory interaction, and semantic assistance. &
        \RaggedRight Coordinates adaptive intervention while retaining the current execution strategy when additional action is unnecessary. \\

        \bottomrule
    \end{tabularx}
\end{table}

The analysis reveals several consistent patterns. First, the adaptive mechanisms are not uniformly activated across missions. The recorded microwave mission in AI2-THOR was completed through the default exploration--detection--navigation--validation sequence without requiring semantic guidance, recovery, or strategy switching. This provides an important inactive-case reference: the adaptive mechanisms do not intervene simply because they are available.

In contrast, more difficult recorded executions exhibit substantially greater adaptive intervention. In the remote-control mission, repeated navigation states and insufficient progress led to strategy switching, recovery activation, semantic guidance, and Mission Memory interaction. The evaluation-wide behavioral measurements show the same general pattern, with higher SGR, SSR, and Recovery Activation values in RoboTHOR and ProcTHOR than in AI2-THOR. These observations are consistent with event-driven intervention during difficult executions; differences in episode length and event prevalence also affect the aggregate indicators.

Mission Memory and candidate validation also exhibit complementary functional roles. Candidate validation determines whether the current observation satisfies the complete target specification, whereas Mission Memory preserves the outcome of earlier candidate and navigation decisions. The recorded laptop mission illustrates this interaction: the first candidate was rejected during validation, its information was stored, exploration continued without repeatedly accepting the same candidate, and a subsequent candidate was later confirmed.

The Semantic Advisor provides another useful contrast between active and inactive conditions. Its activation-to-step ratio remains below 1\% in all three environments, and it is completely absent from straightforward recorded missions that do not require additional contextual reasoning. The low activation-to-step ratio is consistent with conditional semantic guidance. It does not quantify the fraction of episodes using guidance, the number of calls in an affected episode, or the fraction of total inference time and API cost attributable to the Advisor.

Decision Stability is lower in environments with more recorded decision revisions. Because strategy switches and replanning enter its definition, this relationship is partly definitional rather than independent evidence of supervisory quality. A high value can also occur when an ineffective strategy remains unchanged.

Taken together, the execution-trace analysis provides functional evidence that the mechanisms introduced in F2A-ObjectNav are not merely structural elements of the architecture but are engaged under distinct mission conditions and influence the subsequent execution flow. However, because individual mechanisms were not forcibly disabled and re-evaluated under an identical controlled protocol, the analysis should not be interpreted as a quantitative estimate of their independent causal effects on SR or SPL. A removal-based component ablation would be required to establish such isolated performance contributions.

% ----------------------------------------------------------------
\subsection{Indicative Comparison with Related ObjectNav Methods}
% ----------------------------------------------------------------

Table~\ref{tab:related_comparison} places the reported SR and SPL values of F2A-ObjectNav alongside selected ObjectNav methods evaluated on the same named simulation environments. The purpose of this table is to provide numerical context rather than a controlled performance ranking. The cited studies differ in scene subsets, target categories, observation spaces, training procedures, stopping conditions, success definitions, and other evaluation details. Accordingly, differences between the reported values should not be interpreted as evidence of statistically controlled superiority or inferiority between methods.

\begin{table}[!ht]
    \centering
    \caption{Indicative cross-study comparison with selected ObjectNav methods. Values correspond to the protocols reported in the original studies and are not directly controlled across methods.}
    \label{tab:related_comparison}
    \footnotesize
    \ComparisonTableStyle
    \begin{tabularx}{\linewidth}{C{0.16\linewidth}C{0.33\linewidth}*{3}{Z}}
        \toprule
        \rowcolor{ComparisonHeader}
        \textcolor{TableHeaderText}{\textbf{Environment}} &
        \textcolor{TableHeaderText}{\textbf{Method}} &
        \textcolor{TableHeaderText}{\textbf{Year}} &
        \textcolor{TableHeaderText}{\textbf{SR (\%)}} &
        \textcolor{TableHeaderText}{\textbf{SPL}} \\
        \midrule

                \cellcolor{ComparisonStripe} & DAT \cite{dang2023dat} & 2023 & 82.39 & 0.489 \\
        \noalign{\kern-0.2pt}
        \cellcolor{ComparisonStripe} & IOM \cite{xie2023iom} & 2023 & 82.99 & 0.474 \\
        \noalign{\kern-0.2pt}
        \cellcolor{ComparisonStripe} & MT \cite{dang2023mt} & 2023 & 83.14 & 0.502 \\
        \noalign{\kern-0.2pt}
        \cellcolor{ComparisonStripe} & Memory-MambaNav \cite{sun2025memorymamba} & 2025 & 83.19 & 0.498 \\
        \noalign{\kern-0.2pt}
        \cellcolor{ComparisonStripe}\multirow{-5}{=}{\centering AI2-THOR} & F2A-ObjectNav (ours) & 2026 & 84.00 & 0.520 \\

        \midrule

                \cellcolor{ComparisonStripe} & ESC \cite{zhou2023esc} & 2023 & 38.10 & 0.222 \\
        \noalign{\kern-0.2pt}
        \cellcolor{ComparisonStripe} & IOM \cite{xie2023iom} & 2023 & 42.07 & 0.270 \\
        \noalign{\kern-0.2pt}
        \cellcolor{ComparisonStripe} & MT \cite{dang2023mt} & 2023 & 42.08 & 0.290 \\
        \noalign{\kern-0.2pt}
        \cellcolor{ComparisonStripe} & VLTNet \cite{wen2024vltnet} & 2024 & 33.20 & 0.171 \\
        \noalign{\kern-0.2pt}
        \cellcolor{ComparisonStripe} & CogNav \cite{cao2025cognav} & 2025 & 54.60 & 0.243 \\
        \noalign{\kern-0.2pt}
        \cellcolor{ComparisonStripe}\multirow{-6}{=}{\centering RoboTHOR} & F2A-ObjectNav (ours) & 2026 & 40.00 & 0.220 \\

        \midrule

                \cellcolor{ComparisonStripe} & NatVLM (DivScene) \cite{wang2025divscene} & 2025 & 53.12 & 0.444 \\
        \noalign{\kern-0.2pt}
        \cellcolor{ComparisonStripe}\multirow{-2}{=}{\centering ProcTHOR} & F2A-ObjectNav (ours) & 2026 & 44.00 & 0.370 \\

        \bottomrule
    \end{tabularx}
\end{table}

The table documents externally reported values rather than a common-protocol leaderboard. Training requirements and available observations differ across the methods; the present evaluation does not isolate these effects. The published values are retained at a consistent SPL precision, with rounding where necessary.

Because these values originate from independently conducted studies with non-identical experimental protocols, the comparison is used only to position the reported results within the broader ObjectNav literature. The primary contribution of F2A-ObjectNav is therefore not framed as a cross-study performance advantage, but as the introduction and evaluation of an interpretable event-driven coordination architecture for adaptive ObjectNav.

% ================================================================
\section{Discussion}
% ================================================================

The two stages of this study address different but connected questions. Stage I examines whether a fixed combination of exploration, detection, and validation modules can provide a consistently favorable balance among navigation effectiveness, validation reliability, execution time, and model-related cost. The results show that the evaluated configurations occupy different positions in this trade space and that no single fixed configuration is preferable under all considered criteria. Stage II therefore changes the design objective from selecting one configuration before execution to coordinating specialized capabilities according to the evolving mission state.

The Stage II results are consistent with the intended operational behavior of F2A-ObjectNav. In AI2-THOR, semantic guidance, strategy switching, and recovery were activated relatively infrequently, while Decision Stability remained high. RoboTHOR and ProcTHOR produced higher rates of strategy revision, recovery, and semantic intervention under the adopted behavioral metrics. This pattern is consistent with the design principle that additional adaptive mechanisms should be introduced selectively rather than remain continuously active throughout every mission.

The execution-trace analysis provides additional evidence regarding the functional role of these adaptive mechanisms. By combining evaluation-wide behavioral measurements with recorded mission traces, the analysis shows that Mission Memory, recovery, strategy switching, semantic guidance, candidate management, and supervisory coordination become active under different execution conditions and influence the subsequent mission flow. Straightforward episodes can proceed through the default perception--navigation--validation process with little or no adaptive intervention, whereas more difficult executions involve greater use of memory, recovery, strategy revision, and semantic support. At the same time, this evidence remains observational rather than removal-based. Individual mechanisms were not disabled and re-evaluated under an identical matched protocol, and the analysis therefore does not quantify their isolated causal effects on SR or SPL.

The results also emphasize the distinction between final target validation and end-to-end mission completion. Validation Accuracy remained within a relatively narrow range across the three evaluation environments, whereas SR varied substantially. Because VA is candidate-level and SR is episode-level, their difference does not by itself localize a bottleneck. The recorded unsuccessful traces nevertheless illustrate failures both before validation and after repeated candidate rejection. Consequently, future improvements to the architecture should not focus exclusively on the semantic validator, but also on the efficiency and reliability of candidate acquisition and navigation.

The proposed architecture also differs from recent approaches that assign a more direct supervisory role to large language models. For example, CogNav models ObjectNav through multiple cognitive states and uses an LLM as part of the mechanism that determines transitions among those states \cite{cao2025cognav}. F2A-ObjectNav follows a different design principle: mission-level coordination remains rule-based, while language and multimodal models are restricted to specialized functions such as goal interpretation, target validation, and conditional semantic guidance. This separation is intended to preserve explicit supervisory logic and reduce dependence on continuous high-level semantic reasoning. The contribution of F2A-ObjectNav should therefore be interpreted primarily as an architectural approach to coordinating heterogeneous ObjectNav capabilities rather than as a replacement for existing perception, exploration, or navigation methods.

The relationship between Stage I and Stage II is primarily methodological and architectural rather than numerical. The two stages use different evaluation environments and experimental protocols, and their numerical results should therefore not be directly compared. Stage I provides empirical evidence that different fixed pipelines exhibit different performance--efficiency--cost trade-offs, while Stage II uses these observations to motivate mission-level adaptation through supervisory coordination, Mission Memory, candidate management, recovery, strategy revision, and conditional semantic guidance. This fixed-to-adaptive progression constitutes the central design rationale of the present study.

% ================================================================
\section{Limitations}

The complete F2A-ObjectNav architecture was evaluated in simulation over 25 episodes per environment. Each episode therefore contributes four percentage points to the corresponding SR, and the results should be interpreted as descriptive evidence from a limited evaluation rather than a precise estimate of performance across all scenes and targets. Random scene selection does not make this study equivalent to evaluation on the complete official benchmark splits.

The two stages use different environments and protocols, and no matched fixed-versus-adaptive comparison or component-removal ablation is available in Stage II. Execution traces establish that the mechanisms are used during missions, but they do not isolate their causal effects on SR, SPL, or execution cost. Comparisons with independently published methods likewise provide numerical context rather than a controlled ranking.

Behavior depends on the selected models, navigation components, supervisory rules, and activation conditions. The event--response descriptions and prompts explain the architecture, but exact reproduction also depends on implementation-specific thresholds and runtime settings. The behavioral metrics are study-specific diagnostic measures and should be interpreted according to Appendix~\ref{app:metric_definitions}. Task Time and API Cost depend on hardware, service latency, model versions, and pricing, and are not hardware-independent measures of computational efficiency.

Physical deployment would introduce sensor noise, localization errors, actuation uncertainty, dynamic obstacles, and communication constraints not established by the present simulation results. Evaluation under these conditions and a controlled analysis of individual adaptive mechanisms remain necessary to assess broader generalization.

% ================================================================
\section{Conclusion and Future Work}
% ================================================================

This work investigated ObjectNav through a two-stage development process. Stage I examined representative fixed combinations of exploration, object detection, and target validation and revealed clear trade-offs among navigation effectiveness, validation reliability, task time, environmental performance retention, and model-related cost. These findings motivated F2A-ObjectNav, an event-driven adaptive architecture in which a rule-based Supervisor coordinates specialized navigation, perception, validation, memory, recovery, and semantic-reasoning components according to the evolving mission state.

F2A-ObjectNav was evaluated in AI2-THOR, RoboTHOR, and ProcTHOR, achieving SR/SPL values of 84.00\%/0.52, 40.00\%/0.22, and 44.00\%/0.37, respectively. The architecture-specific behavioral measurements showed different patterns of supervisory intervention across the three environments, with higher rates of strategy revision, recovery, and semantic support recorded in RoboTHOR and ProcTHOR than in AI2-THOR.

The execution-trace mechanism analysis further showed that the adaptive mechanisms are engaged differently according to the conditions encountered during execution. Straightforward missions can proceed with little additional intervention, whereas more difficult executions activate different combinations of Mission Memory, recovery, strategy revision, candidate management, and semantic support. This analysis provides functional evidence for the role of the adaptive mechanisms within the integrated architecture. However, because the mechanisms were not independently removed and re-evaluated under a matched protocol, the present results do not establish their isolated causal effects on navigation performance.

The primary contribution of this study is therefore the development of an interpretable fixed-to-adaptive ObjectNav framework in which conventional navigation and perception components remain modular while mission-level events can trigger memory-informed decisions, candidate management, recovery, strategy revision, and conditional semantic guidance. The architecture combines explicit rule-based supervision with specialized perception and semantic components while retaining the ability to execute straightforward missions without unnecessary high-level intervention.

Future work should investigate the complete architecture on a physical mobile robot and conduct controlled removal-based ablation experiments to quantify the individual and combined effects of Mission Memory, recovery, strategy switching, candidate management, and semantic guidance. Further improvements can target exploration efficiency, candidate discovery and approach, supervisory decision rules, and more efficient local or lightweight semantic models to reduce model-related latency and external dependency. Evaluation on larger and more diverse scene sets, as well as more challenging natural-language goals involving attributes, visually similar objects, partial occlusion, and longer navigation horizons, would provide a broader assessment of the architecture's generalization and practical limitations.

% ================================================================
% References
% ================================================================

% Author declarations adapted to the two-author simulation study.
\section*{Declaration of AI use}
The authors used ChatGPT for paraphrasing and language editing of selected manuscript sections and to generate the explanatory schematics in Figures 1 and 5–11. The authors reviewed and edited all AI-assisted content and take responsibility for the manuscript’s accuracy and integrity.

\section*{Statements and Declarations}
\subsection*{Funding}
This study was supported by a grant from Sharif University of Technology.

\subsection*{Conflict of interest}
Seyed Ali Mirghasemi and Alireza Taheri declare that they have no conflict of interest.

\subsection*{Availability of data and materials}
The data and execution records associated with this study are maintained in the archive of the Social and Cognitive Robotics Laboratory, Sharif University of Technology.

\subsection*{Code availability}
The source code developed for the F2A-ObjectNav framework and the reported experiments is maintained in the archive of the Social and Cognitive Robotics Laboratory, Sharif University of Technology.

\subsection*{CRediT authorship contribution statement}
\textbf{Seyed Ali Mirghasemi:} Conceptualization, Methodology, Software, Investigation, Data curation, Formal analysis, Visualization, Writing -- original draft, Writing -- review and editing. \textbf{Alireza Taheri:} Supervision, Writing -- review and editing. Both authors read and approved the final manuscript.

\subsection*{Ethical approval}
This simulation-based study did not involve research on human participants or animals.

\subsection*{Consent for publication}
Not applicable.

\bibliographystyle{unsrt}
\bibliography{references}

% ================================================================
% Appendices
% ================================================================

\clearpage
\appendix

% Appendix heading format:
%   Appendix A
%   Experimental Settings and Evaluation Environments
\titleformat{\section}[display]
  {\normalfont\Large\bfseries\centering}
  {Appendix \thesection}
  {0.35em}
  {\large\bfseries}
\titlespacing*{\section}{0pt}{0pt}{1.4em}

% ----------------------------------------------------------------
\section{Experimental Settings and Evaluation Environments}
\label{app:experimental_settings}
% ----------------------------------------------------------------

This appendix summarizes the principal experimental settings used to evaluate the adaptive F2A-ObjectNav architecture in Stage II. Its purpose is to document the system configuration, model assignments, execution logic, and characteristics of the evaluation environments without repeating the complete architectural description provided in the main text.

The same high-level architecture and supervisory logic are retained across AI2-THOR, RoboTHOR, and ProcTHOR. Simulator-specific differences are handled through the corresponding observation, pose, and action interfaces rather than by modifying the core coordination logic.

\setcounter{table}{0}
\renewcommand{\thetable}{A.\arabic{table}}
\renewcommand{\theHtable}{A\arabic{table}}

\begin{table}[!ht]
    \centering
    \caption{Summary of the principal experimental settings used in the Stage II evaluation.}
    \label{tab:appendixA_settings}
    \scriptsize
    \AppendixATableStyle
    \begin{tabularx}{\linewidth}{C{0.31\linewidth}Z}
        \toprule
        \rowcolor{AppendixAHeader}
        \textcolor{TableHeaderText}{\textbf{Item}} & \textcolor{TableHeaderText}{\textbf{Setting}} \\
        \midrule

        Evaluation type &
        Simulation-based ObjectNav evaluation in indoor environments \\

        Evaluation environments &
        AI2-THOR, RoboTHOR, and ProcTHOR \\

        Mission specification &
        Object category or natural-language target description \\

        Goal representation &
        \RaggedRight Primary object representation for detection together with the complete target description retained for validation \\

        Supervisory controller &
        \RaggedRight Rule-based, event-driven Supervisor; no learned policy or LLM is used as the primary mission-level controller \\

        Agent observations &
        \RaggedRight Visual observations together with available depth, pose, and simulator-specific state information required by the navigation and supervisory components \\

        Agent actions &
        \RaggedRight Simulator-supported translational motion, rotation, viewpoint adjustment, navigation-goal execution, and mission termination \\

        Initial exploration behavior &
        \RaggedRight Frontier-based exploration is used as the default search behavior in the implemented system \\

        Adaptive strategy revision &
        \RaggedRight The Supervisor can revise the active exploration behavior when repeated search, stagnation, or insufficient progress is detected \\

        Goal interpretation model &
        GPT-4o mini \\

        Primary object detector &
        YOLO11x \\

        Open-vocabulary detector &
        YOLO-World v8x \\

        Target validation model &
        GPT-4.1 mini \\

        Semantic guidance model &
        GPT-4o \\

        Semantic-guidance activation &
        \RaggedRight Conditional; requested when conventional search or recovery does not provide sufficient progress or when additional semantic context is required \\

        Candidate handling &
        \RaggedRight Detector outputs are treated as candidates rather than immediate mission success; rejected candidates can be stored in Mission Memory and repeated spatial detections can be consolidated \\

        Mission Memory &
        \RaggedRight Stores mission-relevant episode information such as rejected candidates, repeatedly explored or repeated states, recent goals, recovery-related events, and strategy changes \\

        Recovery behavior &
        \RaggedRight Activated in response to navigation stagnation, repeated trajectories, failed candidate approach, or other detected loss-of-progress conditions \\

        Validation behavior &
        \RaggedRight Candidate observations are evaluated against the complete target description after candidate approach and viewpoint adjustment \\

        Recorded outputs &
        \RaggedRight Execution logs, navigation events, candidate records, validation outcomes, Mission Memory operations, strategy revisions, recovery events, semantic-guidance calls, and evaluation metrics \\

        \bottomrule
    \end{tabularx}
\end{table}

Model roles and activation modes are summarized in Table~\ref{tab:model_roles}; the structured prompts are provided in Appendix~\ref{app:prompts}.

\begin{table}[!ht]
    \centering
    \caption{Roles and main characteristics of the three Stage II evaluation environments.}
    \label{tab:appendixA_environments}
    \footnotesize
    \AppendixATableStyle
    \begin{tabularx}{\linewidth}{C{0.18\linewidth}C{0.27\linewidth}C{0.22\linewidth}Z}
        \toprule
        \rowcolor{AppendixAHeader}
        \textcolor{TableHeaderText}{\textbf{Environment}} &
        \textcolor{TableHeaderText}{\textbf{Role in the evaluation}} &
        \textcolor{TableHeaderText}{\textbf{Agent type}} &
        \textcolor{TableHeaderText}{\textbf{Main characteristic}} \\
        \midrule

        AI2-THOR &
        Evaluation in structured residential scenes &
        Virtual first-person agent &
        \RaggedRight Interactive indoor scenes including kitchens, living rooms, bedrooms, and bathrooms \\

        RoboTHOR &
        Evaluation under robot-oriented navigation conditions &
        LoCoBot-based simulated agent &
        \RaggedRight Indoor navigation with motion and perception constraints designed around a mobile-robot platform \\

        ProcTHOR &
        Evaluation under greater structural and layout diversity &
        Virtual agent in procedurally generated environments &
        \RaggedRight Procedurally generated houses with varied layouts, room connectivity, and object arrangements \\

        \bottomrule
    \end{tabularx}
\end{table}

The environments are used as distinct simulation settings, not as matched difficulty levels. Table~\ref{tab:episode_manifest} summarizes the common episode protocol.

\begin{table}[!ht]
\centering
\caption{Summary of the Stage II episode protocol.}
\label{tab:episode_manifest}
\footnotesize
\AppendixATableStyle
\begin{tabularx}{\linewidth}{C{0.24\linewidth}Z}
\toprule
\rowcolor{AppendixAHeader}
\textcolor{TableHeaderText}{\textbf{Setting}} & \textcolor{TableHeaderText}{\textbf{Specification}} \\
\midrule
Evaluation size & 25 episodes per environment; 75 in total \\
Scene selection & \RaggedRight Random selection from the available scenes of each environment \\
Target specification & \RaggedRight Object category or natural-language description; target present, location withheld \\
Initialization & \RaggedRight Valid starting pose; mission state and Mission Memory reset per episode \\
Agent inputs & RGB observations, depth, and navigation-related pose/state information \\
Target interpretation & \RaggedRight LLM-based category extraction and preparation of a detailed validation description \\
Execution steps & \RaggedRight Recorded operations, including movement, rotation, detection, and recovery \\
Termination & Final candidate acceptance or execution-budget exhaustion \\
Success conditions & \RaggedRight Stopping with the correct target visible and within 1~m of the camera; correctness checked by human inspection \\
Incorrect acceptance & Episode scored as unsuccessful \\
\bottomrule
\end{tabularx}
\end{table}

Model roles are listed in Table~\ref{tab:model_roles}, and prompts are provided in Appendix~\ref{app:prompts}. Task Time is averaged over all 25 episodes of each environment, including unsuccessful episodes, and includes motion, perception, model inference, and API waiting during mission execution. API Cost denotes the reported external-model expenditure per episode; it excludes local computation and hardware costs.

\clearpage

% ----------------------------------------------------------------
\section{Definitions and Calculation of Evaluation Metrics}
\label{app:metric_definitions}
% ----------------------------------------------------------------
\begingroup
\small
\setlength{\abovedisplayskip}{5pt plus 1pt minus 1pt}
\setlength{\belowdisplayskip}{5pt plus 1pt minus 1pt}
\setlength{\abovedisplayshortskip}{4pt plus 1pt minus 1pt}
\setlength{\belowdisplayshortskip}{4pt plus 1pt minus 1pt}
\titlespacing*{\subsection}{0pt}{0.85em}{0.35em}
\titlespacing*{\paragraph}{0pt}{0.45em}{0.35em}

This appendix presents the mathematical definitions and calculation procedures for the metrics used throughout the study. To distinguish their roles clearly, the measures are organized into four groups: standard ObjectNav metrics, system-level performance measures, a Stage-I-specific environmental-retention measure, and architecture-specific behavioral metrics.

The standard ObjectNav metrics are Success Rate (SR) and Success-weighted Path Length (SPL). Task Time (TT), Validation Accuracy (VA), and API Cost (AC) are treated as system-level measures. Environmental Adaptability Score (EAS) is used only in Stage I to summarize the retention of navigation performance from the simple to the complex layout. The remaining metrics characterize the internal behavior of the adaptive F2A-ObjectNav architecture and are diagnostic measures introduced for this study rather than established ObjectNav benchmark metrics.

\setcounter{equation}{0}
\renewcommand{\theequation}{B.\arabic{equation}}
\renewcommand{\theHequation}{B\arabic{equation}}

\setcounter{table}{0}
\renewcommand{\thetable}{B.\arabic{table}}
\renewcommand{\theHtable}{B\arabic{table}}

% ----------------------------------------------------------------
\subsection{Standard ObjectNav Metrics}
% ----------------------------------------------------------------

\paragraph{Success Rate.}

Success Rate (SR) is the proportion of evaluated missions that satisfy the success criterion used in the corresponding experimental protocol:

\begin{equation}
    \mathrm{SR}
    =
    \frac{N_{\mathrm{success}}}
         {N_{\mathrm{total}}}.
    \label{eq:sr}
\end{equation}

Here, $N_{\mathrm{success}}$ is the number of successfully completed missions and $N_{\mathrm{total}}$ is the total number of evaluated missions. The same episode-level success definition used during the corresponding benchmark evaluation is applied when calculating SR.

\paragraph{Success-weighted Path Length.}

Success-weighted Path Length (SPL) evaluates mission success together with path efficiency using the standard embodied-navigation formulation \cite{anderson2018evaluation}:

\begin{equation}
    \mathrm{SPL}
    =
    \frac{1}{N}
    \sum_{i=1}^{N}
    S_i
    \frac{L_i^{*}}
         {\max\!\left(L_i,L_i^{*}\right)}.
    \label{eq:spl}
\end{equation}

Here, $S_i \in \{0,1\}$ indicates whether episode $i$ satisfies the adopted success criterion, $L_i^{*}$ is the shortest feasible path length to the target under the corresponding evaluation setting, $L_i$ is the path length traveled by the agent, and $N$ is the number of evaluated episodes.

A higher SPL indicates that a larger proportion of missions are completed successfully while following paths closer to the corresponding shortest feasible route. Path lengths refer to spatial distances rather than counts of recorded execution operations.

% ----------------------------------------------------------------
\subsection{System-Level Performance Measures}
% ----------------------------------------------------------------

\paragraph{Task Time.}

Task Time (TT) is the average elapsed wall-clock time from mission initialization to termination:

\begin{equation}
    \mathrm{TT}
    =
    \frac{1}{N}
    \sum_{i=1}^{N}
    \left(
    t_i^{\mathrm{end}}
    -
    t_i^{\mathrm{start}}
    \right).
    \label{eq:tt}
\end{equation}

Here, $t_i^{\mathrm{start}}$ and $t_i^{\mathrm{end}}$ denote the recorded start and termination times of episode $i$. The measured duration includes robot or agent motion as well as delays introduced by perception, model inference, target validation, semantic guidance, supervisory decisions, and other execution operations.

\paragraph{Validation Accuracy.}

Validation Accuracy (VA) measures the correctness of candidate-validation decisions:

\begin{equation}
    \mathrm{VA}
    =
    \frac{TP+TN}
         {TP+FP+TN+FN},
    \label{eq:va}
\end{equation}

where $TP$ denotes valid target candidates correctly accepted, $FP$ denotes non-target candidates incorrectly accepted, $TN$ denotes non-target candidates correctly rejected, and $FN$ denotes valid target candidates incorrectly rejected.

Candidate decisions were checked by human inspection against the requested object identity and any specified attributes. Acceptance of an incorrect candidate and rejection of a correct candidate were treated as validation errors.

VA therefore characterizes the semantic validation component rather than overall navigation success. A mission may fail before a candidate reaches the validation stage, and a high VA should consequently not be interpreted as equivalent to a high SR.

\paragraph{API Cost.}

For components that access externally hosted language, vision-language, or multimodal models, the average model-related inference cost per episode is defined as

\begin{equation}
    \mathrm{AC}
    =
    \frac{1}{N}
    \sum_{i=1}^{N}
    c_i^{\mathrm{API}},
    \label{eq:api_cost}
\end{equation}

Here, $c_i^{\mathrm{API}}$ is the total external-model inference cost for episode $i$, obtained by summing all billable usage categories at their applicable rates during the original runs, and $N$ is the number of evaluated episodes. Billable categories may include input tokens, output tokens, image processing, or model invocations, depending on the provider's pricing structure. API Cost excludes local inference, hardware, and energy costs. Comparisons depend on the billing categories and prices in effect during the original runs.

% ----------------------------------------------------------------
\subsection{Stage-I-Specific Environmental Adaptability Score}
% ----------------------------------------------------------------

\paragraph{Environmental Adaptability Score.}

Environmental Adaptability Score (EAS) is a Stage-I-specific diagnostic measure used to summarize how well a fixed configuration retains navigation performance when moving from the simple layout to the complex layout. It is defined as

\begin{equation}
    \mathrm{EAS}
    =
    \frac{1}{2}
    \left(
    \frac{\mathrm{SR}_{\mathrm{complex}}}
         {\mathrm{SR}_{\mathrm{simple}}}
    +
    \frac{\mathrm{SPL}_{\mathrm{complex}}}
         {\mathrm{SPL}_{\mathrm{simple}}}
    \right).
    \label{eq:eas}
\end{equation}

EAS therefore averages the relative retention of success rate and path efficiency under increased environmental complexity. A value of 1 denotes equal average relative performance across the two layouts, and values below 1 indicate average relative degradation. The score is not bounded above by 1: it can exceed 1 if complex-layout performance improves. It measures relative retention, not absolute task performance or the presence of an adaptive mechanism.

EAS is not a standard ObjectNav benchmark metric and is used only for the paired simple-versus-complex comparison performed in Stage I. Equation~\ref{eq:eas} is defined when both $\mathrm{SR}_{\mathrm{simple}}$ and $\mathrm{SPL}_{\mathrm{simple}}$ are nonzero.

Using the SR and SPL values reported in the Stage I results table, the corresponding EAS values are approximately 0.273, 0.593, 0.595, 0.790, 0.759, and 0.890 for C1--C6, respectively. Small differences may occur if EAS is recalculated from unrounded run-level values rather than the rounded SR and SPL values displayed in the table.

% ----------------------------------------------------------------
\subsection{Architecture-Specific Behavioral Metrics}
% ----------------------------------------------------------------

The following metrics characterize the internal behavior of F2A-ObjectNav. They quantify the frequency or relative involvement of semantic guidance, strategy revision, recovery, Mission Memory, and high-level decision changes during mission execution.

These measures are intended as architecture-specific diagnostic indicators. They complement SR, SPL, and the system-level measures but should not be interpreted independently as measures of navigation quality or causal component effectiveness.

Unless otherwise stated, the evaluation-level value is calculated as the arithmetic mean of the corresponding episode-level quantity. This means that each episode contributes equally to the final reported value, irrespective of episode length.

\paragraph{Semantic Guidance Rate.}

Semantic Guidance Rate (SGR) measures the frequency with which the Semantic Advisor is activated relative to the number of recorded episode steps:

\begin{equation}
    \mathrm{SGR}
    =
    \frac{1}{N}
    \sum_{i=1}^{N}
    \frac{C_i^{\mathrm{sem}}}
         {\max(1,T_i)}.
    \label{eq:sgr}
\end{equation}

Here, $C_i^{\mathrm{sem}}$ is the number of Semantic Advisor activations in episode $i$, and $T_i$ is the corresponding number of recorded episode steps.

A low SGR is not inherently undesirable. In the proposed architecture, semantic guidance is intentionally designed as a conditional resource and may remain unused in straightforward episodes.

\paragraph{Strategy Switching Rate.}

Strategy Switching Rate (SSR) measures the frequency with which the active navigation or exploration strategy is revised:

\begin{equation}
    \mathrm{SSR}
    =
    \frac{1}{N}
    \sum_{i=1}^{N}
    \frac{S_i^{\mathrm{sw}}}
         {\max(1,T_i)},
    \label{eq:ssr}
\end{equation}

where $S_i^{\mathrm{sw}}$ is the number of recorded navigation or exploration strategy changes in episode $i$.

A higher SSR indicates that the supervisory layer revised the active search behavior more frequently relative to episode length. It does not by itself indicate whether the revisions improved mission outcome.

\paragraph{Recovery Activation.}

Recovery Activation (RA) measures the degree to which detected navigation problems are accompanied by recovery actions:

\begin{equation}
    \mathrm{RA}
    =
    \frac{1}{N}
    \sum_{i=1}^{N}
    \begin{cases}
        \min\!\left(
        1,
        \dfrac{R_i^{\mathrm{ev}}}
              {N_i^{\mathrm{issue}}}
        \right),
        & N_i^{\mathrm{issue}}>0,\\[6pt]
        0,
        & N_i^{\mathrm{issue}}=0.
    \end{cases}
    \label{eq:recovery_activation}
\end{equation}

Here, $R_i^{\mathrm{ev}}$ is the number of recovery actions executed in episode $i$, and $N_i^{\mathrm{issue}}$ is the number of detected navigation problems. These problems can include stagnation, repeated trajectories, failure to reach a candidate, unproductive rotation, or prolonged lack of useful progress.

The value is capped at 1 for each episode to prevent multiple recovery actions associated with the same or overlapping navigation difficulties from producing values above the intended range.

Episodes with no detected navigation issue contribute a value of zero. Consequently, the evaluation-level RA reflects both the prevalence of navigation problems and the system's recovery response to those problems. It should therefore not be interpreted as a pure recovery-success probability.

\paragraph{Memory Usage (Memory-Assisted Loop Response).}

Memory Usage (MU), operationalized here as Memory-Assisted Loop Response, measures the extent to which Mission Memory contributes to responses associated with detected repeated or unproductive navigation states:

\begin{equation}
    \mathrm{MU}
    =
    \frac{1}{N}
    \sum_{i=1}^{N}
    \begin{cases}
        \min\!\left(
        1,
        \dfrac{M_i^{\mathrm{esc}}}
              {N_i^{\mathrm{loop}}}
        \right),
        & N_i^{\mathrm{loop}}>0,\\[6pt]
        0,
        & N_i^{\mathrm{loop}}=0.
    \end{cases}
    \label{eq:memory_usage}
\end{equation}

Here, $M_i^{\mathrm{esc}}$ is the number of memory-supported escape or corrective actions in episode $i$, and $N_i^{\mathrm{loop}}$ is the number of detected trajectory-repetition or navigation-loop events.

This metric does not represent the total number of Mission Memory reads or writes. It specifically quantifies memory-supported responses associated with detected loop or repetition events.

As with Recovery Activation, episodes without a detected loop contribute zero to the evaluation-level average. The reported value therefore reflects both the occurrence of loop events and the involvement of Mission Memory in responding to them.

\paragraph{Decision Stability.}

Decision Stability (DS) measures the consistency of high-level system decisions relative to episode length. Strategy switches, detection-procedure switches, and replanning operations are treated as high-level decision revisions:

\begin{equation}
    \mathrm{DS}
    =
    \frac{1}{N}
    \sum_{i=1}^{N}
    \frac{1}{
    1+
    \dfrac{
    S_i^{\mathrm{sw}}
    +
    D_i^{\mathrm{sw}}
    +
    R_i^{\mathrm{pl}}
    }
    {\max(1,T_i)}
    }.
    \label{eq:decision_stability}
\end{equation}

Here, $S_i^{\mathrm{sw}}$ is the number of strategy switches, $D_i^{\mathrm{sw}}$ is the number of recorded detection-model or detection-procedure switches, $R_i^{\mathrm{pl}}$ is the number of path-replanning operations, and $T_i$ is the number of recorded episode steps.

Higher DS values correspond to fewer high-level revisions relative to episode length. However, Decision Stability is not a measure of mission success: an episode can remain highly stable while following an ineffective strategy. DS should therefore always be interpreted together with SR, SPL, and the remaining behavioral metrics.

The preceding definitions specify the aggregation and zero-event conventions used for interpreting the reported behavioral metrics.

For readability, SGR, SSR, Recovery Activation, Memory-Assisted Loop Response, and Decision Stability are multiplied by 100 and reported as percentages in the main Results section.

% ----------------------------------------------------------------
\subsection{Summary of Symbols}
% ----------------------------------------------------------------

\begin{table}[!ht]
    \centering
    \caption{Symbols used in the architecture-specific behavioral metrics.}
    \label{tab:appendixB_symbols}
    \scriptsize
    \AppendixBTableStyle
    \begin{tabularx}{\linewidth}{C{0.16\linewidth}C{0.29\linewidth}Z}
        \toprule
        \rowcolor{AppendixBHeader}
        \textcolor{TableHeaderText}{\textbf{Symbol}} &
        \textcolor{TableHeaderText}{\textbf{Term}} &
        \textcolor{TableHeaderText}{\textbf{Description}} \\
        \midrule

        $C_i^{\mathrm{sem}}$ &
        Semantic Guidance Call Count &
        Number of Semantic Advisor activations in episode $i$. \\

        $S_i^{\mathrm{sw}}$ &
        Strategy Switch Count &
        \RaggedRight Number of navigation or exploration strategy changes in episode $i$. \\

        $R_i^{\mathrm{ev}}$ &
        Recovery Event Count &
        Number of recovery actions executed in episode $i$. \\

        $N_i^{\mathrm{issue}}$ &
        Navigation Issue Count &
        \RaggedRight Number of detected navigation problems in episode $i$, including stagnation, repeated trajectories, candidate-approach failure, unproductive rotation, or prolonged lack of progress. \\

        $M_i^{\mathrm{esc}}$ &
        Memory-Supported Escape Count &
        \RaggedRight Number of corrective or escape actions supported by Mission Memory in episode $i$. \\

        $N_i^{\mathrm{loop}}$ &
        Navigation Loop Count &
        \RaggedRight Number of detected trajectory-repetition or navigation-loop events in episode $i$. \\

        $D_i^{\mathrm{sw}}$ &
        Detection Switch Count &
        \RaggedRight Number of recorded changes between detection models or detection procedures in episode $i$. \\

        $R_i^{\mathrm{pl}}$ &
        Replanning Count &
        Number of path-replanning operations in episode $i$. \\

        $T_i$ &
        Episode Step Count &
        \RaggedRight Number of recorded episode steps used as the normalization basis for the corresponding behavioral metrics. \\

        $N$ &
        Number of Episodes &
        \RaggedRight Total number of episodes included in the evaluation-level average. \\

        \bottomrule
    \end{tabularx}
\end{table}

The standard and system-level metrics are reported either in their native range or as percentages according to the presentation used in the main Results tables. The architecture-specific behavioral metrics are reported as percentages after multiplication by 100. EAS is reported as a dimensionless Stage-I-specific performance-retention score.
\endgroup

\clearpage

% ----------------------------------------------------------------
\section{Prompts and Textual Inputs Used in the System}
\label{app:prompts}
% ----------------------------------------------------------------
\begingroup
\setlength{\parskip}{0pt}
\titlespacing*{\subsection}{0pt}{0.9em}{0.35em}

This appendix documents the structured prompts and controlled textual inputs used by the language and multimodal components of F2A-ObjectNav. These inputs support three principal functions: mission-goal interpretation, conditional semantic guidance, and final candidate validation. The prompts are designed around specialized component roles and structured outputs rather than unrestricted conversational interaction.

YOLO11x and YOLO-World v8x are included for completeness, although they do not receive free-form prompts in the same manner as the GPT-based components. YOLO11x operates on its supported detection classes together with the alias mapping maintained by the system, whereas YOLO-World v8x receives a controlled list of target classes and equivalent names.

\setcounter{table}{0}
\renewcommand{\thetable}{C.\arabic{table}}
\renewcommand{\theHtable}{C\arabic{table}}

\begin{table}[!ht]
    \centering
    \caption{Summary of the prompts and controlled textual inputs used in F2A-ObjectNav.}
    \label{tab:appendixC_prompt_summary}
    \scriptsize
    \AppendixCTableStyle
    \begin{tabularx}{\linewidth}{C{0.23\linewidth}C{0.31\linewidth}C{0.17\linewidth}Z}
        \toprule
        \rowcolor{AppendixCHeader}
        \textcolor{TableHeaderText}{\textbf{Component}} &
        \textcolor{TableHeaderText}{\textbf{Role}} &
        \textcolor{TableHeaderText}{\textbf{Model}} &
        \textcolor{TableHeaderText}{\textbf{Activation}} \\
        \midrule

        Target Interpreter &
        \RaggedRight Converts the mission goal into a detector-facing object class and a complete validation query &
        GPT-4o mini &
        Mission initialization \\

        Semantic Advisor &
        \RaggedRight Suggests a promising direction or region for continuing the search &
        GPT-4o &
        \RaggedRight Conditional; used after insufficient progress, repeated navigation, or unsuccessful recovery \\

        Attribute-Aware Vision Validator &
        \RaggedRight Validates a candidate against the complete target description and explicit attributes &
        GPT-4.1 mini &
        Attribute-constrained target descriptions \\

        Alias-Based Vision Validator &
        \RaggedRight Validates a candidate using the target name and accepted equivalent terms &
        GPT-4.1 mini &
        Simple target categories without additional descriptive constraints \\

        YOLO-World Class List &
        \RaggedRight Defines the controlled target classes used for open-vocabulary detection &
        YOLO-World v8x &
        Open-vocabulary detection \\

        YOLO11x &
        Performs primary detection for supported object categories &
        YOLO11x &
        Routine object detection \\

        \bottomrule
    \end{tabularx}
\end{table}

% ----------------------------------------------------------------
\subsection{Mission Goal Interpretation Prompt}
% ----------------------------------------------------------------

The mission-goal interpretation prompt is used once at the beginning of each episode. Its purpose is to separate the detector-facing object category from descriptive constraints that should be retained for later semantic validation.

For example, if the user requests a \emph{white refrigerator}, the primary detection class is reduced to \emph{refrigerator}, while the complete phrase \emph{white refrigerator} is preserved as the validation query. This separation prevents descriptive attributes from unnecessarily constraining object detection while preserving the complete user intent for final candidate assessment.

\Needspace{22\baselineskip}
The structured prompt template is:

\begin{promptbox}
You parse object-goal navigation targets for an embodied agent.
Given the user's target description, return STRICT JSON only with keys:
{"object_name":"<short canonical object class for detection>",
 "validation_query":"<full user intent including attributes/constraints>",
 "reason":"<short>"}

Rules:
- object_name must be a short lowercase class label suitable for object detectors
  (examples: refrigerator, microwave, oven, mug, apple).
- validation_query must preserve ALL user constraints
  (color, material, size, subtype, or location hints when explicitly provided).
- If the user says "white refrigerator",
  object_name = "refrigerator" and
  validation_query must preserve "white refrigerator".
- Do not invent constraints not present in the user text.

User target description: {raw_input}
\end{promptbox}

\Needspace{8\baselineskip}
A representative structured output is:

\begin{outputbox}
{
  "object_name": "refrigerator",
  "validation_query": "white refrigerator",
  "reason": "The object category is refrigerator and the color constraint must be preserved."
}
\end{outputbox}

Equivalent object names or aliases used by the detection and validation components are maintained separately by the system and are not required to be generated by the Target Interpreter.

% ----------------------------------------------------------------
\subsection{Semantic Guidance Prompt}
% ----------------------------------------------------------------

The Semantic Advisor is activated only when additional high-level contextual support is required. Typical activation conditions include prolonged lack of progress, repeated exploration, unsuccessful recovery, or failure to identify a useful candidate.

The model receives the current target together with a summary of recent search progress and multiple candidate viewing directions. Its role is to identify a promising direction or indicate that the current region is unlikely to contain useful target-related context.

The returned information is advisory and does not directly generate low-level motion commands. The Supervisor interprets the recommendation, and the Navigation and Exploration Agent converts the resulting high-level decision into an executable navigation objective.

\Needspace{22\baselineskip}
The structured prompt template is:

\begin{promptbox}
You are a semantic navigator advisor for object navigation.

Target object: {target_name}
Steps since progress: {steps_since_progress}

Given multi-direction views, select the most promising direction
for continuing the search.

Use object co-occurrence and scene-context information when useful
(example: a microwave may be associated with a kitchen,
counter, refrigerator, or nearby appliances).

If all directions appear irrelevant to the target context,
set leave_room=true.

Return STRICT JSON only with this schema:
{"best_view_index": <int>,
 "confidence": <0..1>,
 "leave_room": <true|false>,
 "reason": "<short>"}

Views metadata: {views_summary}
\end{promptbox}

\Needspace{8\baselineskip}
A representative structured output is:

\begin{outputbox}
{
  "best_view_index": 2,
  "confidence": 0.82,
  "leave_room": false,
  "reason": "This view contains furniture and objects that are contextually associated with the target."
}
\end{outputbox}

% ----------------------------------------------------------------
\subsection{Attribute-Aware Target Validation Prompt}
% ----------------------------------------------------------------

When the target contains explicit descriptive constraints, such as color, material, size, or subtype, candidate validation is performed against the complete target specification rather than only the general object category.

The validator receives the candidate observation together with the complete validation query and determines whether the visible object satisfies all explicitly requested constraints. The examples below illustrate the interface and do not constitute a separate quantitative evaluation of attribute-constrained navigation.

\Needspace{22\baselineskip}
The structured prompt template is:

\begin{promptbox}
You are validating object-goal navigation with attribute-aware matching.

User target specification: {query}

Accept ONLY if the visible object satisfies the full user specification,
including attributes such as color, material, size, or subtype
when these are explicitly specified.

Reject near-miss objects
(example: black refrigerator when the user requested white refrigerator).

Return STRICT JSON only:
{"is_target_present": <true|false>,
 "confidence": <0..1>,
 "reason": "<short>"}
\end{promptbox}

\Needspace{8\baselineskip}
A representative structured output is:

\begin{outputbox}
{
  "is_target_present": true,
  "confidence": 0.91,
  "reason": "The visible refrigerator is white and matches the complete target description."
}
\end{outputbox}

The validator therefore returns a binary target-presence decision together with a confidence value and a short explanation. If the visual evidence available to the system is insufficient for reliable assessment, the surrounding navigation and supervisory logic may obtain an additional observation or adjust the viewpoint before another validation attempt. This behavior should not be interpreted as a third textual output class of the validator.

% ----------------------------------------------------------------
\subsection{Alias-Based Target Validation Prompt}
% ----------------------------------------------------------------

For target categories without additional descriptive constraints, validation can use equivalent object names maintained by the system. The validator receives the primary target name together with the accepted alias list and determines whether the intended object is visible in the candidate observation.

\Needspace{22\baselineskip}
The structured prompt template is:

\begin{promptbox}
You are validating object-goal navigation.

Target object aliases are: {aliases_txt}

Treat the supplied aliases as equivalent names for the intended target.
Accept the candidate if the intended target object is visibly present
under any of these equivalent names.

Return STRICT JSON only:
{"is_target_present": <true|false>,
 "confidence": <0..1>,
 "reason": "<short>"}
\end{promptbox}

\Needspace{8\baselineskip}
A representative structured output is:

\begin{outputbox}
{
  "is_target_present": false,
  "confidence": 0.87,
  "reason": "The visible object is a bowl and does not match the target or its accepted aliases."
}
\end{outputbox}

As in attribute-aware validation, the returned decision is binary. Viewpoint adjustment or collection of additional evidence is handled by the surrounding architecture when required rather than represented as a separate validator output state.

% ----------------------------------------------------------------
\subsection{YOLO-World Textual Input}
% ----------------------------------------------------------------

YOLO11x does not receive a free-form natural-language prompt. It operates on its supported object classes, while equivalent terms and aliases are handled by the surrounding system logic.

YOLO-World v8x receives a controlled list of target classes derived from the primary object representation and accepted equivalent terms. \Needspace{6\baselineskip}
The class list is supplied programmatically:

\begin{codebox}
model.set_classes(list(labels))
\end{codebox}

\Needspace{5\baselineskip}
For example, when the mission target is \emph{mug}, the controlled class list may be:

\begin{codebox}
["mug", "cup"]
\end{codebox}

Accordingly, the textual input supplied to YOLO-World v8x is a constrained detection vocabulary rather than an unrestricted natural-language prompt.

% ----------------------------------------------------------------
\subsection{Prompt-Design Considerations}
% ----------------------------------------------------------------

The prompts used in F2A-ObjectNav are intentionally task-specific and structured. Each language or multimodal component receives only the mission information required for its assigned function and is instructed to return a predefined JSON structure that can be interpreted by the supervisory logic.

The system does not rely on a single general-purpose language-model prompt for mission control. Instead, goal interpretation, semantic guidance, and candidate validation are separated into distinct interfaces. This design reduces ambiguity in inter-component communication and supports the modular organization of the architecture.

The prompts reported in this appendix should therefore be interpreted as component interfaces rather than as autonomous navigation policies. Low-level motion, exploration, candidate approach, recovery, and mission-level coordination remain outside the direct control of the language and multimodal prompts.
\endgroup

\clearpage

% ----------------------------------------------------------------
\section{Execution Evidence and Representative Mission Traces}
\label{app:execution_evidence}
% ----------------------------------------------------------------

This appendix presents selected execution records generated during evaluation of the complete F2A-ObjectNav architecture. The examples are intended to provide qualitative execution evidence for the internal behavior discussed in the main Results section rather than to serve as additional evaluation-level performance measurements.

The examples were selected to illustrate contrasting execution patterns, not sampled to estimate their prevalence. The selected missions illustrate four distinct execution patterns: direct successful completion with minimal adaptive intervention, successful rejected-candidate handling using Mission Memory, unsuccessful execution involving repeated adaptive intervention, and unsuccessful execution in which candidate acquisition remained the dominant limitation.

These examples also provide the episode-level evidence used in the execution-trace mechanism analysis presented in the main text.

\setcounter{table}{0}
\renewcommand{\thetable}{D.\arabic{table}}
\renewcommand{\theHtable}{D\arabic{table}}

\setcounter{figure}{0}
\renewcommand{\thefigure}{D.\arabic{figure}}
\renewcommand{\theHfigure}{D\arabic{figure}}

% ----------------------------------------------------------------
\subsection{Structure of the Recorded Outputs}
% ----------------------------------------------------------------

Several execution artifacts are stored for each evaluated episode to support later inspection of mission behavior. These records include general run metadata, structured event sequences, component handoffs, validation evidence, Mission Memory interactions, exploration maps, and complete episode videos.

Table~\ref{tab:appendixD_outputs} summarizes the main recorded outputs.

\begin{table}[!ht]
    \centering
    \caption{Principal outputs recorded during F2A-ObjectNav execution.}
    \label{tab:appendixD_outputs}
    \footnotesize
    \AppendixDTableStyle
    \begin{tabularx}{\linewidth}{C{0.30\linewidth}Z}
        \toprule
        \rowcolor{AppendixDHeader}
        \textcolor{TableHeaderText}{\textbf{Output}} & \textcolor{TableHeaderText}{\textbf{Role in the analysis}} \\
        \midrule

        \texttt{run\_meta.json} &
        \RaggedRight Stores episode-level metadata such as environment, target, final status, recorded step count, execution time, and summary information. \\

        \texttt{flow\_compact.json} &
        \RaggedRight Provides a compact structured sequence of the principal execution events and state transitions. \\

        \texttt{flow\_compact.txt} &
        \RaggedRight Provides a human-readable summary of the mission execution flow. \\

        \texttt{appendix\_bundle.json} &
        \RaggedRight Contains execution information organized for qualitative analysis and appendix reporting. \\

        \texttt{appendix\_handoffs.txt} &
        \RaggedRight Records selected coordination events and information exchanges among system components. \\

        \texttt{exploration\_map.png} &
        \RaggedRight Visualizes explored regions together with the trajectory recorded during the episode. \\

        \texttt{validation\_crop\_step\_*.jpg} &
        Stores cropped candidate observations used during target validation. \\

        \texttt{validation\_frame\_step\_*.jpg} &
        \RaggedRight Stores the corresponding full-scene observation at a validation step. \\

        \texttt{run.mp4} &
        Contains the complete video recording of the episode. \\

        \bottomrule
    \end{tabularx}
\end{table}

The execution traces are used here to illustrate when adaptive mechanisms become active and how their activation changes the subsequent mission flow. They should not be interpreted as substitutes for the aggregate evaluation-level metrics reported in the main Results section.

% ----------------------------------------------------------------
\subsection{Representative Execution Examples}
% ----------------------------------------------------------------

% ----------------------------------------------------------------
\subsubsection{Direct Successful Execution with Minimal Intervention}
% ----------------------------------------------------------------

The first example is a successful AI2-THOR mission with \emph{microwave} as the target. The system detected a candidate, navigated toward it, obtained an appropriate observation, and confirmed the same candidate during final validation.

\begin{table}[!ht]
    \centering
    \caption{Summary of the direct successful execution for the microwave target.}
    \label{tab:appendixD_microwave}
    \footnotesize
    \AppendixDTableStyle
    \begin{tabularx}{0.86\linewidth}{C{0.38\linewidth}Z}
        \toprule
        \rowcolor{AppendixDHeader}
        \textcolor{TableHeaderText}{\textbf{Property}} & \textcolor{TableHeaderText}{\textbf{Value}} \\
        \midrule

        Environment & AI2-THOR \\
        Scene & FloorPlan1 \\
        Target & Microwave \\
        Final status & Successful \\
        Recorded steps & Approximately 43 \\
        Execution time & Approximately 42 s \\
        Initial exploration strategy & Frontier-based exploration \\
        Detection model & YOLO-World v8x \\
        Validation model & GPT-4.1 mini \\
        Candidate status & One candidate detected and confirmed \\
        Semantic guidance & Not required \\
        Recovery & Not required \\
        Strategy switching & Not required \\

        \bottomrule
    \end{tabularx}
\end{table}

\paragraph{Mission execution flow.}

\begin{enumerate}
    \item Mission initialization and collection of bootstrap information.
    \item Activation of frontier-based exploration, YOLO-World v8x detection, and GPT-4.1 mini validation.
    \item Exploration of the environment.
    \item Detection of a microwave candidate.
    \item Navigation toward the candidate and adjustment of the observation viewpoint.
    \item Candidate validation using GPT-4.1 mini.
    \item Confirmation of the target and successful mission termination.
\end{enumerate}

This episode illustrates the default execution path of F2A-ObjectNav. Because useful progress was maintained and the detected candidate was successfully validated, no additional semantic guidance, recovery, or strategy revision was required.

This case is important for interpreting the adaptive architecture: the availability of adaptive mechanisms does not imply that they are continuously active. Straightforward missions can be completed using the default perception--navigation--validation sequence alone.

\begin{figure}[H]
    \centering
    \IfFileExists{figures/figD1.pdf}{
        \includegraphics[width=0.5\linewidth,keepaspectratio]{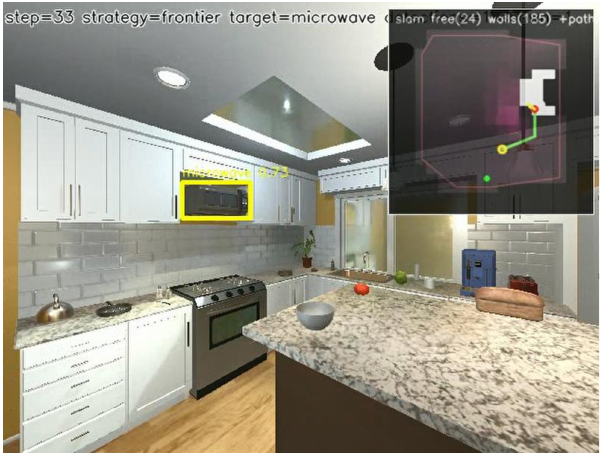}
    }{
        \fbox{\parbox[c][4.8cm][c]{0.90\linewidth}{
            \centering
            \textbf{Figure placeholder}\\[4pt]
            Direct successful microwave-target mission in AI2-THOR\\[4pt]
            \texttt{figures/figD1.pdf}
        }}
    }
    \caption{Representative direct successful execution for the microwave target in AI2-THOR.}
    \label{fig:appendixD_microwave}
\end{figure}

% ----------------------------------------------------------------
\subsubsection{Successful Execution with Rejected-Candidate Management}
% ----------------------------------------------------------------

The second example is a successful AI2-THOR mission with \emph{laptop} as the target. Unlike the previous example, the first detected candidate was not accepted during validation. Its information was recorded in Mission Memory, exploration continued, and a later candidate was subsequently confirmed. This recorded behavior directly demonstrates the separation between candidate detection and final target acceptance.

\begin{table}[!ht]
    \centering
    \caption{Summary of the successful execution involving rejected-candidate management.}
    \label{tab:appendixD_laptop}
    \footnotesize
    \AppendixDTableStyle
    \begin{tabularx}{0.86\linewidth}{C{0.38\linewidth}Z}
        \toprule
        \rowcolor{AppendixDHeader}
        \textcolor{TableHeaderText}{\textbf{Property}} & \textcolor{TableHeaderText}{\textbf{Value}} \\
        \midrule

        Environment & AI2-THOR \\
        Scene & FloorPlan201 \\
        Target & Laptop \\
        Final status & Successful \\
        Recorded steps & Approximately 63 \\
        Execution time & Approximately 67 s \\
        Initial exploration strategy & Frontier-based exploration \\
        Detection model & YOLO-World v8x \\
        Validation model & GPT-4.1 mini \\
        Candidate status & One initial candidate rejected; a later candidate confirmed \\
        Mission Memory role & Stored the rejected candidate for subsequent candidate filtering \\
        Subsequent action & Exploration resumed after candidate rejection \\

        \bottomrule
    \end{tabularx}
\end{table}

\paragraph{Mission execution flow.}

\begin{enumerate}
    \item Mission initialization and bootstrap execution.
    \item Activation of frontier-based exploration, YOLO-World v8x detection, and GPT-4.1 mini validation.
    \item Detection and approach of the first laptop candidate.
    \item Validation of the candidate using GPT-4.1 mini.
    \item Rejection of the initial candidate.
    \item Storage of the rejected candidate in Mission Memory.
    \item Resumption of exploration while accounting for the previous candidate record.
    \item Detection and approach of another candidate.
    \item Collection of a suitable viewpoint and repeated validation.
    \item Confirmation of the laptop target and successful mission termination.
\end{enumerate}

This example demonstrates the interaction among object detection, target validation, and Mission Memory. Detector output is treated as a candidate rather than immediate mission success. When validation rejects a candidate, the outcome can be retained by Mission Memory and used to influence later execution.

The mission therefore provides direct qualitative evidence for the functional role assigned to candidate management and episode-level memory in F2A-ObjectNav.

\begin{figure}[H]
    \centering
    \IfFileExists{figures/figD2.pdf}{
        \includegraphics[width=0.5\linewidth,keepaspectratio]{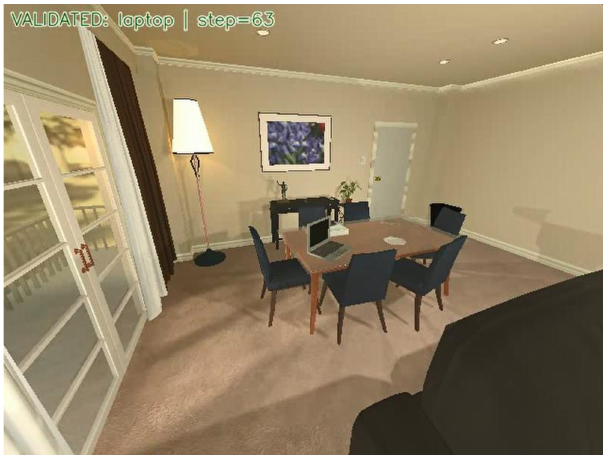}
    }{
        \fbox{\parbox[c][4.8cm][c]{0.90\linewidth}{
            \centering
            \textbf{Figure placeholder}\\[4pt]
            Laptop-target mission with rejected-candidate handling\\[4pt]
            \texttt{figures/figD2.pdf}
        }}
    }
    \caption{Successful laptop-target mission after rejection and storage of an earlier candidate.}
    \label{fig:appendixD_laptop}
\end{figure}

% ----------------------------------------------------------------
\subsubsection{Unsuccessful Execution with Multiple Adaptive Interventions}
% ----------------------------------------------------------------

The third example is an unsuccessful AI2-THOR mission with \emph{remote control} as the target. It was selected because the recorded episode contains several forms of adaptive intervention, including repeated-state detection, strategy switching, recovery, semantic guidance, candidate validation, and Mission Memory interaction.

\begin{table}[!ht]
    \centering
    \caption{Summary of the unsuccessful remote-control mission involving multiple adaptive mechanisms.}
    \label{tab:appendixD_remote}
    \scriptsize
    \AppendixDTableStyle
    \begin{tabularx}{0.86\linewidth}{C{0.38\linewidth}Z}
        \toprule
        \rowcolor{AppendixDHeader}
        \textcolor{TableHeaderText}{\textbf{Property}} & \textcolor{TableHeaderText}{\textbf{Value}} \\
        \midrule

        Environment & AI2-THOR \\
        Scene & FloorPlan205 \\
        Target & Remote control \\
        Final status & Unsuccessful \\
        Recorded steps & Approximately 1,000 \\
        Execution time & Approximately 253 s \\
        Initial exploration strategy & Frontier-based exploration \\
        Detection model & YOLO-World v8x \\
        Validation model & GPT-4.1 mini \\
        Candidate status & Several candidates examined; none finally confirmed \\
        Semantic guidance & Activated multiple times \\
        Strategy switching & Recorded between frontier-based and random exploration \\
        Recovery & Activated multiple times during unproductive navigation states \\
        Mission Memory role & Stored rejected candidates and supported repetition management \\
        Final outcome & No valid target candidate was confirmed before termination \\

        \bottomrule
    \end{tabularx}
\end{table}

\paragraph{Mission execution flow.}

\begin{enumerate}
    \item Mission initialization and bootstrap execution.
    \item Activation of frontier-based exploration, YOLO-World v8x detection, and GPT-4.1 mini validation.
    \item Occurrence of repeated states and navigation difficulties during exploration.
    \item Strategy revision between frontier-based and random exploration.
    \item Activation of recovery operations, including actions such as \texttt{rotate\_scan} and \texttt{relocalize}.
    \item Conditional activation of semantic guidance.
    \item Detection and inspection of multiple potential target candidates.
    \item Rejection of invalid candidates during validation.
    \item Storage of rejected-candidate information in Mission Memory.
    \item Continued exploration until the recorded episode execution limit was reached.
    \item Unsuccessful termination without confirmation of a valid target.
\end{enumerate}

This example shows that unsuccessful mission completion does not imply that the adaptive mechanisms remained inactive. The Supervisor repeatedly modified the execution process through recovery, strategy revision, semantic assistance, and memory-supported candidate handling.

At the same time, these interventions did not produce successful target confirmation within the available execution horizon. The episode therefore illustrates an important limitation of adaptive coordination: activating additional mechanisms can modify mission behavior and attempt to restore progress, but it does not guarantee successful task completion.

This distinction is also important for interpreting the execution-trace analysis. The presence of recovery, semantic guidance, or strategy switching provides evidence that the mechanisms are functionally active, but their activation alone cannot be interpreted as evidence of a positive causal effect on SR.

\begin{figure}[H]
    \centering
    \IfFileExists{figures/figD3.pdf}{
        \includegraphics[width=0.5\linewidth,keepaspectratio]{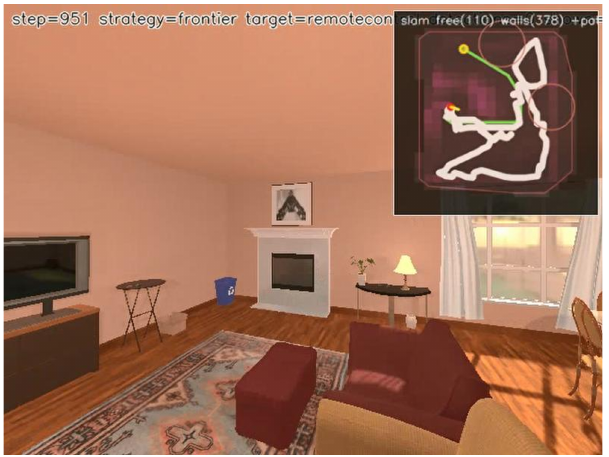}
    }{
        \fbox{\parbox[c][4.8cm][c]{0.90\linewidth}{
            \centering
            \textbf{Figure placeholder}\\[4pt]
            Unsuccessful remote-control mission with adaptive interventions\\[4pt]
            \texttt{figures/figD3.pdf}
        }}
    }
    \caption{Snapshot during an unsuccessful remote-control mission involving strategy revision, recovery, semantic guidance, and Mission Memory; the displayed step precedes episode termination.}
    \label{fig:appendixD_remote}
\end{figure}

% ----------------------------------------------------------------
\subsubsection{Unsuccessful Execution Dominated by Candidate-Acquisition Failure}
% ----------------------------------------------------------------

The final example is an unsuccessful ProcTHOR mission with \emph{soccer ball} as the target. In contrast to the preceding remote-control mission, the main recorded difficulty was candidate acquisition rather than repeated rejection of detected candidates. No valid target candidate was reached during the available execution horizon.

\begin{table}[!ht]
    \centering
    \caption{Summary of the unsuccessful soccer-ball mission in ProcTHOR.}
    \label{tab:appendixD_soccer}
    \scriptsize
    \AppendixDTableStyle
    \begin{tabularx}{0.86\linewidth}{C{0.38\linewidth}Z}
        \toprule
        \rowcolor{AppendixDHeader}
        \textcolor{TableHeaderText}{\textbf{Property}} & \textcolor{TableHeaderText}{\textbf{Value}} \\
        \midrule

        Environment & ProcTHOR \\
        Scene & ProcTHOR-train-00057 \\
        Target & Soccer ball \\
        Final status & Unsuccessful \\
        Recorded steps & Approximately 1,000 \\
        Execution time & Approximately 405 s \\
        Initial exploration strategy & Frontier-based exploration \\
        Primary detection model & YOLO11x \\
        Validation model & GPT-4.1 mini \\
        Candidate status & No valid candidate reached for final validation \\
        Semantic guidance & Activated multiple times \\
        Strategy switching & Activated multiple times \\
        Recovery & Activated multiple times \\
        Mission Memory role & Supported responses to some repeated navigation states \\
        Final outcome & No valid target candidate was reached before termination \\

        \bottomrule
    \end{tabularx}
\end{table}

\paragraph{Mission execution flow.}

\begin{enumerate}
    \item Mission initialization and bootstrap execution.
    \item Activation of frontier-based exploration, YOLO11x detection, and GPT-4.1 mini validation.
    \item Exploration in the procedurally generated ProcTHOR environment.
    \item Failure to obtain a valid soccer-ball candidate for approach and validation.
    \item Occurrence of repeated states and navigation difficulties.
    \item Strategy revision between frontier-based and random exploration.
    \item Activation of recovery at multiple stages.
    \item Conditional use of semantic guidance to support continued search.
    \item Use of Mission Memory during responses to repeated navigation states.
    \item Continued search until the recorded episode execution limit was reached.
    \item Unsuccessful termination because no valid target candidate became available for final validation.
\end{enumerate}

This episode illustrates a different failure mode from the remote-control example. Here, the principal bottleneck occurred before target validation: the system did not acquire and approach a valid candidate within the available execution horizon.

Recovery, strategy revision, semantic guidance, and Mission Memory were nevertheless activated during the search. The example therefore shows that adaptive intervention cannot fully compensate for failures in candidate discovery and acquisition when suitable perceptual evidence is not obtained.

This trace identifies a failure before final validation and illustrates why candidate discovery and approach must be considered separately from candidate-validation accuracy.

\begin{figure}[H]
    \centering
    \IfFileExists{figures/figD4.pdf}{
        \includegraphics[width=0.5\linewidth,keepaspectratio]{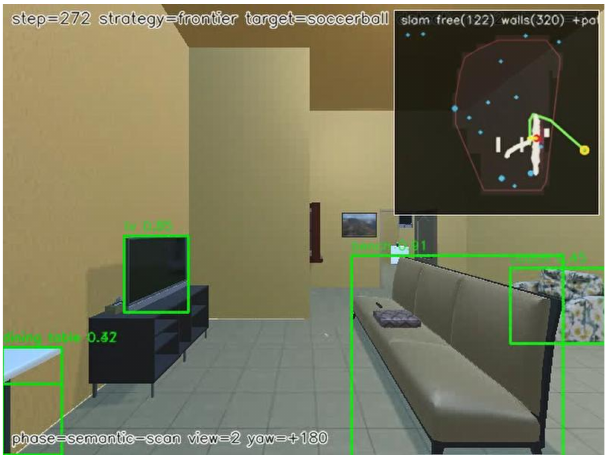}
    }{
        \fbox{\parbox[c][4.8cm][c]{0.90\linewidth}{
            \centering
            \textbf{Figure placeholder}\\[4pt]
            Unsuccessful soccer-ball mission in ProcTHOR\\[4pt]
            \texttt{figures/figD4.pdf}
        }}
    }
    \caption{Snapshot during an unsuccessful ProcTHOR mission in which no valid soccer-ball candidate was reached for final validation; the displayed step precedes episode termination.}
    \label{fig:appendixD_soccer}
\end{figure}

% ----------------------------------------------------------------
\subsection{Relation to the Execution-Trace Mechanism Analysis}
% ----------------------------------------------------------------

The four traces illustrate low-intervention success, successful rejection and subsequent candidate discovery, repeated adaptive intervention without success, and failure before final validation. They show how mechanisms participate in mission execution, but neither their activation nor these selected examples establishes an isolated causal improvement in SR or SPL. This evidence complements the aggregate results and is not a substitute for a controlled component-removal ablation.

% Author biographies follow all appendices.
\clearpage
\begingroup
\newcommand{\authorphoto}[1]{%
  \IfFileExists{#1}{\includegraphics[width=30mm,height=40mm,keepaspectratio]{#1}}%
  {\fbox{\parbox[c][38mm][c]{28mm}{\centering Author\\photograph}}}%
}

\titleformat{\section}{\normalfont\Large\bfseries}{\thesection}{1em}{}
\section*{Author biographies}
\setlength{\parindent}{0pt}
\begin{minipage}[t]{0.22\textwidth}
\vspace{0pt}\authorphoto{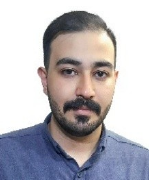}
\end{minipage}\hfill
\begin{minipage}[t]{0.74\textwidth}
\vspace{0pt}Seyed Ali Mirghasemi received the B.Sc. degree in mechanical engineering from the University of Tehran, Iran, in 2023, and the M.Sc. degree in mechatronics engineering from Sharif University of Technology, Iran, in 2026, ranking first in his program. He is currently a Research Assistant with the Social Robotics Laboratory at Sharif University of Technology. His research interests include autonomous robotics and navigation, embodied AI, robot perception, vision-language models, and LLM-based intelligent systems, including LLM-Agent-Controller.
\end{minipage}

\vspace{12mm}
\begin{minipage}[t]{0.22\textwidth}
\vspace{0pt}\authorphoto{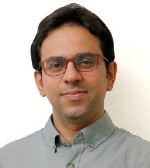}
\end{minipage}\hfill
\begin{minipage}[t]{0.74\textwidth}
\vspace{0pt}Alireza Taheri is an Associate Professor of Mechanical Engineering at Sharif University of Technology, Iran, specializing in social and cognitive robotics. He heads the Social and Cognitive Robotics Laboratory and the Measurement Systems Laboratory. His research focuses on robotics, virtual reality, and human-robot interaction for the education and rehabilitation of children with special needs. His interests include robot design, serious games, artificial intelligence and control, cognitive architectures, and behavioral modeling for adaptive human-robot interaction.
\end{minipage}

\endgroup
\end{document}